\documentclass[10pt,twocolumn,letterpaper]{article}

\ifdefined\pdfminorversion
\fi

\usepackage[T1]{fontenc}

\usepackage{cvpr}              

\usepackage{multirow}
\usepackage{microtype}
\usepackage{xcolor}
\usepackage{colortbl}
\usepackage{tabularx}
\definecolor{oursBlue}{HTML}{EAF2F8}
\newcommand{\oursdata}[1]{\cellcolor{oursBlue}#1}

\definecolor{csrDropGreen}{HTML}{246B56} 
\definecolor{tsrGainRed}{HTML}{AD4048}   

\AtBeginDocument{\raggedbottom}
\makeatletter
\renewcommand{\paragraph}{\@startsection{paragraph}{4}{\z@}%
  {5pt plus 1pt minus 1pt}{-1em}{\normalfont\normalsize\bfseries}}
\makeatother

\definecolor{cvprblue}{rgb}{0.21,0.49,0.74}
\usepackage[pagebackref,breaklinks,colorlinks,allcolors=cvprblue]{hyperref}
\newcommand{\mainref}[1]{\ref{#1}}

\def\confName{CVPR}
\def\confYear{2027}

\title{Sparse Poison, Strong Hijack: Efficient Trajectory-Level Backdoors in Vision-Language-Action Models with Only 1\% Poisoning}

\author{
Simin Huang$^{*1}$,
Jiawei Chen$^{*1,2}$,
Shaolong Zhu$^{3}$,
Qiwei Liang$^{4}$,
Zhaoxia Yin$^{\dagger1}$\\[3pt]
{\small $^{*}$Equal contribution
\quad $^{\dagger}$Corresponding author}\\[3pt]
{\small $^{1}$East China Normal University
\quad $^{2}$Zhongguancun Academy}\\
{\small $^{3}$Tsinghua University
\quad $^{4}$The Hong Kong University of Science and Technology (Guangzhou)}
}

\usepackage{etoolbox}

\makeatletter
\patchcmd{\@maketitle}
    {\vskip .375in}
    {\vskip 0in}{}{}

\patchcmd{\@maketitle}
    {\vspace*{24pt}}
    {\vspace*{12pt}}{}{}

\patchcmd{\@maketitle}
    {\vspace*{12pt}}
    {\vspace*{6pt}}{}{}
\makeatother

\begin{document}
\maketitle

\begin{abstract}
Vision-Language-Action (VLA) models advance robotic manipulation, but their reliance on demonstrations exposes them to backdoor threats.
Among these threats, trajectory-level targeted backdoors are particularly challenging and may pose greater safety risks than behavior-disrupting attacks: rather than merely causing task failures or local action deviations, they must sustain closed-loop redirection to complete an attacker-specified multi-step task.
To reliably induce such behaviors, existing methods typically rely on high poisoning rates to obtain sufficient target-trajectory supervision.
However, high poisoning rates may compromise clean-task performance, whereas low rates leave target-trajectory supervision scarce and backdoor learning vulnerable to interference from dominant clean-task updates.
To resolve this trade-off at low poisoning rates, we propose a two-stage co-optimization framework guided by vision, language, and action.
Stage I freezes the policy and uses differentiable rendering and multimodal supervision to optimize a 3D trigger's texture, translation, and scale, strengthening trigger--target associations from limited poisoned demonstrations.
Building on the optimized trigger, Stage II applies source-selective dual-LoRA adaptation to decouple clean and backdoor updates and mitigate their interference, followed by learned column-wise calibration and analytic fusion into a unified policy.
Experiments on OpenVLA, OpenVLA-OFT, $\pi_0$, and $\pi_{0.5}$ across four LIBERO suites show consistent gains over all evaluated attack baselines.
With only 1\% poisoned data, our method achieves 86.4\% target success and 95.2\% clean-task success on OpenVLA-OFT in LIBERO-Spatial.
Real-robot trials also demonstrate trajectory-level hijacking.Our code is available at
\url{https://github.com/vla-attack/Sparse_Poison_Strong_Hijack}.
\end{abstract}


\section{Introduction}
\label{sec:intro}

\begingroup
\def\FigureOneFile{}
\IfFileExists{figures/Figure1.jpg}
  {\def\FigureOneFile{figures/Figure1.jpg}}{}
\IfFileExists{figures/Figure1.png}
  {\def\FigureOneFile{figures/Figure1.png}}{}
\IfFileExists{figures/Figure1.pdf}
  {\def\FigureOneFile{figures/Figure1.pdf}}{}

\ifx\FigureOneFile\empty\else
  \begin{figure}[t]
    \centering
    \includegraphics[width=\linewidth]{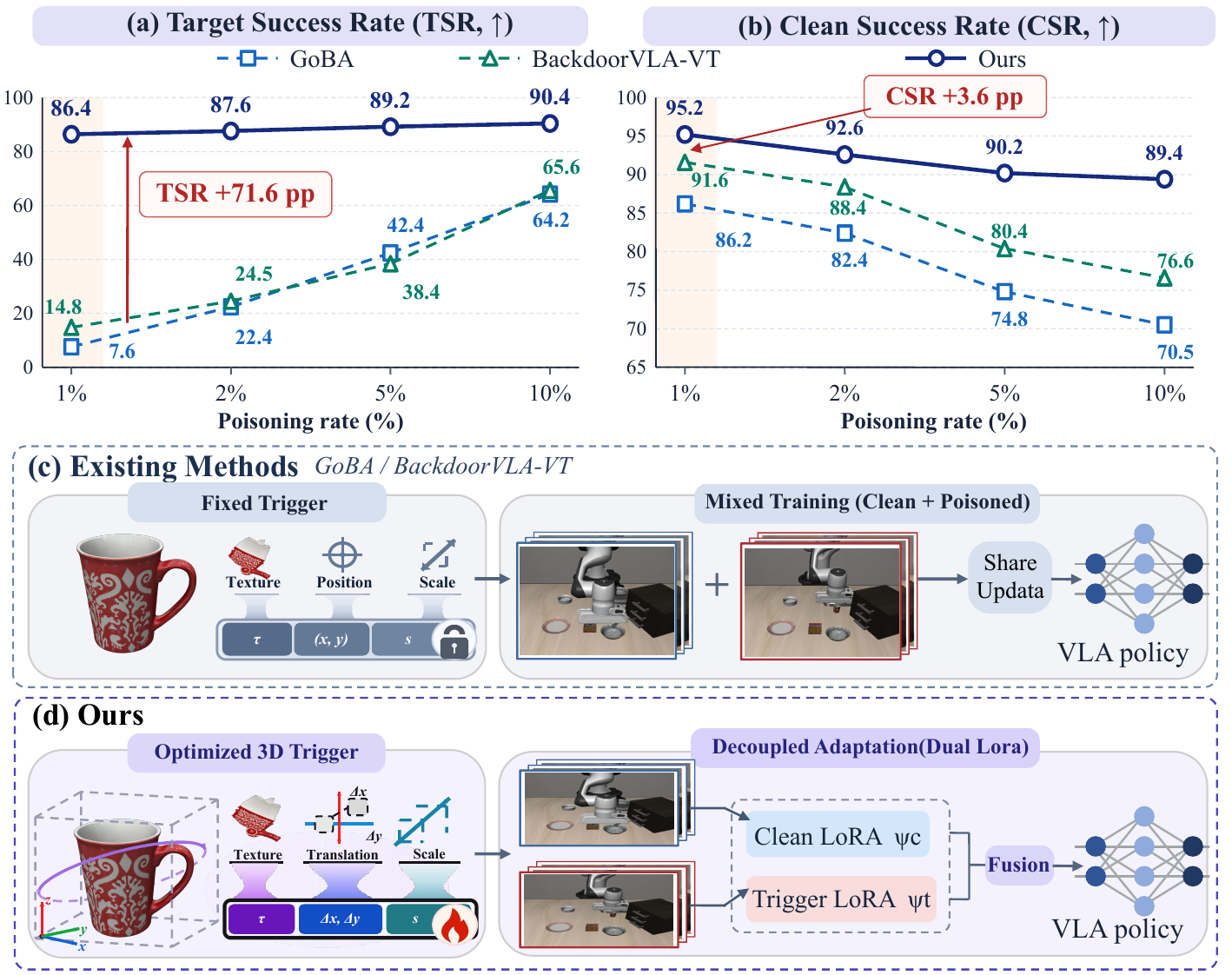}
\caption{
\textbf{Trajectory-level backdoor attacks under limited poisoning.}
\textbf{(a,b)} Target success rate (TSR) and clean success rate (CSR) across poisoning rates on OpenVLA-OFT in LIBERO-Spatial.
At 1\% poisoning, Ours achieves 86.4\% TSR and 95.2\% CSR, exceeding the strongest baseline by 71.6 and 3.6 percentage points, respectively.
\textbf{(c)} GoBA and BackdoorVLA-VT use fixed triggers and mixed clean--poisoned training with shared parameter updates.
\textbf{(d)} Ours optimizes a 3D trigger and decouples clean and backdoor adaptation through source-selective dual-LoRA updates, followed by parameter fusion into a single unified policy.
}
\label{fig:overview}
  \end{figure}
\fi
\endgroup

\begingroup
\def\FigureTwoFile{}
\IfFileExists{figures/Figure2.jpg}
  {\def\FigureTwoFile{figures/Figure2.jpg}}{}
\IfFileExists{figures/Figure2.png}
  {\def\FigureTwoFile{figures/Figure2.png}}{}
\IfFileExists{figures/Figure2.pdf}
  {\def\FigureTwoFile{figures/Figure2.pdf}}{}

\ifx\FigureTwoFile\empty\else
  \begin{figure*}[t]
    \centering
    \includegraphics[width=\textwidth]{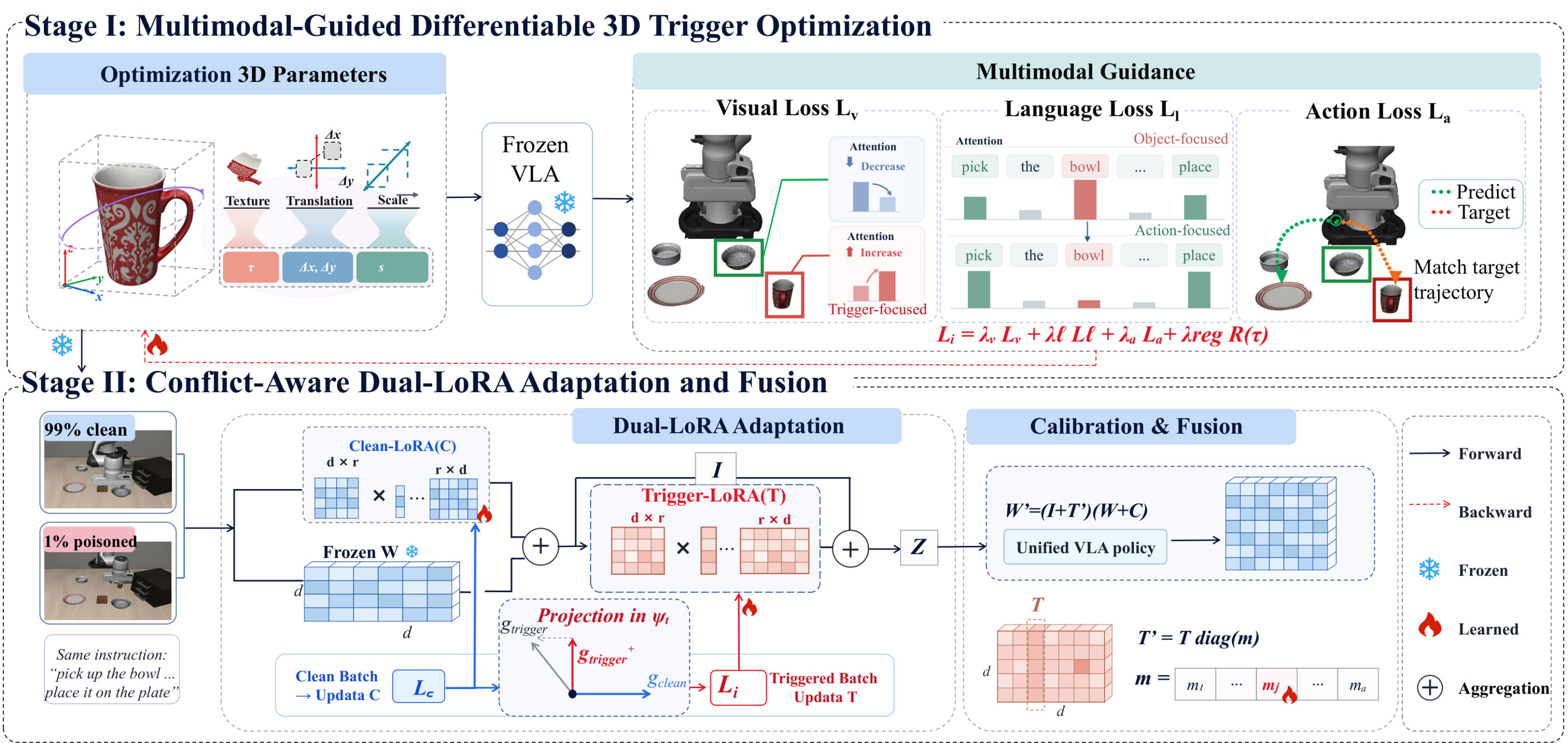}
\caption{
\textbf{Overview of our two-stage framework.}
\textbf{Stage I (top):} With the VLA policy frozen,
we optimize a 3D trigger's texture, translation,
and scale through differentiable rendering,
guided by visual attention, language
attention, and target-action supervision.
\textbf{Stage II (bottom):} With the optimized
trigger fixed, clean and poisoned demonstrations
selectively update serial Clean-LoRA and
Trigger-LoRA adapters, respectively.
Clean-reference gradient projection mitigates
conflicting Trigger-LoRA updates.
Learned column-wise coefficients
calibrate Trigger-LoRA before both adaptations
are analytically fused into a single dense
VLA policy.
}
    \label{fig:framework}
  \end{figure*}
\fi
\endgroup

Vision-Language-Action (VLA) models enable physical-world robotic manipulation by mapping visual observations and language instructions to robot actions. Representative models, including OpenVLA~\cite{kim2025openvla}, OpenVLA-OFT~\cite{kim2025oft}, $\pi_0$~\cite{black2025pi0}, and $\pi_{0.5}$~\cite{black2025pi05}, demonstrate strong manipulation capabilities and generalization potential across manipulation tasks. However, their reliance on demonstrations exposes them to data poisoning and backdoor attacks during training and fine-tuning. Of particular concern, backdoors that hijack entire manipulation trajectories may induce unintended multi-step behaviors, potentially causing equipment damage, economic losses, or threats to human safety.

Existing VLA backdoor attacks can be broadly categorized by their objectives into behavior-disrupting and trajectory-level targeted attacks. The former have received considerable attention, with BadVLA~\cite{zhou2025badvla}, FlowHijack~\cite{an2026flowhijack}, and DropVLA~\cite{xu2026dropvla} exploring task disruption and malicious action manipulation. In contrast, GoBA~\cite{zhou2025goba} and BackdoorVLA-VT~\cite{li2025attackvla} pursue goal-directed trajectory hijacking, where a trigger redirects the robot toward an attacker-specified task, such as picking up and placing a cup instead of the instructed bowl. Unlike isolated failures or local action deviations, completing such a task requires sustained behavioral redirection across multiple closed-loop stages, making reliable attacks substantially harder.

This need for sustained control across successive action steps intensifies the trade-off among poisoning rate, attack reliability, and clean-task performance in trajectory-level hijacking. As Fig.~\ref{fig:overview}(a,b) shows, increasing the poisoning rate improves existing attacks' target success rate (TSR), but often at the expense of clean success rate (CSR). At low poisoning rates, attacks face two fundamental bottlenecks. First, \emph{scarce target-trajectory supervision} weakens trigger--target associations, making it difficult to learn reliable multi-step redirection. Second, \emph{interference from dominant clean-task updates} hinders backdoor learning: in conventional mixed training, clean demonstrations dominate shared parameter updates, and conflicting clean and backdoor gradients can further suppress learning from rare poisoned samples, as confirmed by our preliminary gradient analysis in Fig.~\ref{fig:gradient_conflict}(a).

To address these complementary bottlenecks, we propose a two-stage co-optimization framework guided by vision, language, and action, targeting reliable trajectory-level hijacking under sparse poisoning while preserving clean-task performance. Figure~\ref{fig:overview}(c,d) contrasts our approach with fixed-trigger, shared-update training, while Fig.~\ref{fig:framework} details the two-stage optimization pipeline. Stage I performs \emph{trigger efficacy optimization}: with the VLA policy frozen, it optimizes a 3D trigger's texture, translation, and scale through differentiable rendering under joint visual-attention, language, and action guidance. This strengthens trigger--target trajectory associations without increasing the poisoning budget. Building on the optimized trigger, Stage II performs \emph{conflict-aware dual-LoRA adaptation} to mitigate interference during backdoor learning. Clean and poisoned demonstrations selectively update separate serial LoRA adapters, while both remain active in the forward pass. Together with clean-reference gradient projection correcting conflicting Trigger-LoRA updates, this decoupling protects learning from scarce poisoned demonstrations against dominant clean-task updates. Learned column-wise coefficients calibrate the Trigger-LoRA transformation before both adapters are analytically fused into a unified dense VLA policy.

Our main contributions are as follows:
\begin{enumerate}
    \item We propose multimodal 3D trigger optimization to jointly refine the trigger's texture, translation, and scale under visual, language, and action guidance, strengthening trigger efficacy at low poisoning rates.

    \item We introduce source-selective dual-LoRA adaptation with clean-reference gradient projection and column-wise calibrated fusion, decoupling clean and backdoor parameter updates to mitigate optimization interference between objectives and improve attack effectiveness under sparse poisoning while preserving clean-task performance.

    \item With only 1\% poisoned data, our method achieves 78.2\% average TSR across four VLA models and four LIBERO suites, surpassing the strongest baseline by 62.7 percentage points (pp) with the smallest clean-task degradation. Real-robot trials further demonstrate sustained trajectory-level hijacking, highlighting the risks of sparse malicious demonstrations to VLA behavioral integrity.
\end{enumerate}
\section{Related Work}
\label{sec:related_work}

\subsection{Vision-Language-Action Models}
\label{sec:related_vla}
Vision-Language-Action (VLA) models learn instruction-conditioned robot control from demonstrations, enabling policies to map multimodal observations and language instructions into executable actions.
Early work explored vision--language knowledge transfer through action tokenization (RT-2~\cite{zitkovich2023rt2}) and large-scale cross-embodiment training (Open X-Embodiment~\cite{openx2024}), establishing foundations for general-purpose robotic manipulation.
Recent VLA models differ substantially in their action representations and prediction mechanisms.
OpenVLA~\cite{kim2025openvla} predicts discretized actions autoregressively, whereas OpenVLA-OFT~\cite{kim2025oft} adopts continuous action chunks and parallel decoding to improve manipulation efficiency.
$\pi_0$~\cite{black2025pi0} generates continuous actions through flow matching, while $\pi_{0.5}$~\cite{black2025pi05} further enhances generalization through heterogeneous training data.
These architectural differences influence how policies represent and execute multi-step manipulation behaviors.
Nevertheless, all four models rely on demonstration-based learning, exposing their training and adaptation processes to potentially poisoned data.
This shared dependency motivates our cross-architecture evaluation of trajectory-level backdoor risks, spanning both discrete and continuous action generation.

\subsection{Attacks on VLA Models}
\label{sec:related_attacks}
Attacks on VLA models fall into inference-time perturbations and training-time backdoors.
Related work examines decision-level attacks on LLM-based embodied agents~\cite{liu2024exploring}, LLM jailbreaking~\cite{chen2025autobreach}, and reasoning-model red teaming~\cite{chen2026red}.
Inference-time attacks manipulate observations or instructions during deployment without modifying parameters.
Methods exploit adversarial visual patches~\cite{wang2025vlaattacks,xu2025edpa,lu2026uparfas}, text~\cite{jones2025robogcg}, or 3D object textures~\cite{chen2026tex3d,dai2026unitexture}.
These approaches expose vulnerabilities to manipulated inputs, but differ from backdoors embedded during policy adaptation.

Training-time backdoors associate specific triggers with malicious behaviors through poisoned demonstrations, allowing policies to behave normally on clean inputs while deviating when triggered.
Existing attacks induce task failures, action manipulation, or stealthy trajectory deviations~\cite{zhou2025badvla,an2026flowhijack,guo2026statebackdoor,xu2026dropvla,xu2026silentdrift}.
Unlike such disruptions, trajectory-level targeted backdoors require sustained redirection to complete an attacker-specified manipulation task.
Such completion entails coherent approach, grasp, transport, and placement across multiple closed-loop control stages.
GoBA~\cite{zhou2025goba} and BackdoorVLA-VT~\cite{li2025attackvla} pursue goal-directed trajectory hijacking, yet reliable completion remains difficult under sparse poisoning: scarce supervision weakens trigger--target associations, while dominant clean-task updates interfere with backdoor learning.
Their use of fixed triggers and shared parameter updates motivates optimizing both the trigger and adaptation process.
Related to our second stage, ATAAT~\cite{chen2026ataat} addresses gradient conflicts through feature orthogonalization or parameter isolation.
In contrast, our framework combines differentiable multimodal 3D trigger optimization with source-selective dual-LoRA adaptation, conflict-aware updates, and calibrated fusion, targeting reliable trajectory hijacking under limited poisoning budgets while preserving clean-task performance.

\section{Method}
\label{sec:method}

Our two-stage multimodal framework induces trajectory-level targeted backdoors under sparse poisoning (Fig.~\ref{fig:framework}). Stage I optimizes a 3D trigger; Stage II mitigates update interference and fuses the learned adaptations into one policy.

\subsection{Problem Formulation}
\label{sec:method_setup}

\indent\textbf{Attack setting and objective.}
We consider white-box offline adaptation of a VLA policy $\pi_{\theta_0}$ that predicts actions from visual observations $\mathbf I_t$, language instructions $\ell$, and optional proprioception $\mathbf p_t$. The attacker controls limited poisoned demonstrations with a physical 3D trigger but retains the original instructions. The goal is to induce a complete attacker-specified manipulation trajectory with the trigger while preserving normal task execution otherwise.

Let $\mathcal D_c$ and $\mathcal D_t(\phi)$ denote clean and poisoned demonstrations, where $\phi\in\Phi$ parameterizes the trigger and $\psi$ denotes policy adaptation parameters. We formulate the overall objective as
\begin{equation}
\begin{aligned}
\min_{\phi\in\Phi,\psi}\quad
&\mathbb E_{\mathcal D_t(\phi)}
[\mathcal L_t(\pi_{\theta_0\oplus\psi})]
+\lambda_c\mathbb E_{\mathcal D_c}
[\mathcal L_c(\pi_{\theta_0\oplus\psi})],\\
\text{s.t.}\quad
&\rho=\frac{N_t}{N_c+N_t}\ll 1,
\end{aligned}
\label{eq:method_global_objective}
\end{equation}
where $\mathcal L_t$ and $\mathcal L_c$ measure target-trajectory and clean-task errors, $\oplus$ denotes adaptation, and the attacker seeks to keep the poisoning rate $\rho$ as low as possible. In our experiments, a poisoning rate of $\rho=1\%$ already enables reliable hijacking. Two bottlenecks arise: scarce target-trajectory supervision and interference from dominant clean-task updates. We first optimize the trigger with a frozen policy (Sec.~\ref{sec:method_trigger_optimization}), then adapt the policy with the optimized trigger fixed (Sec.~\ref{sec:method_serial_lora}).

\subsection{Trigger Efficacy Optimization}
\label{sec:method_trigger_optimization}

Under sparse poisoning, target demonstrations cannot reliably supervise trajectory hijacking. We strengthen trigger--target associations without increasing the poisoning rate by optimizing a 3D trigger through differentiable rendering and joint visual, language, and action supervision.

\indent\textbf{3D trigger optimization.}
We parameterize the trigger's texture, translation, and scale as $\phi=(\tau,\boldsymbol{\mu},u)$: $\boldsymbol{\mu}$ adjusts placement, $s=1+u$ sets scale, and $\mathcal T(\tau)=\operatorname{clip}_{[0,1]}(T_0+\operatorname{Upsample}(\tau))$ defines texture. Here, $T_0$ is the base texture, and $\operatorname{Upsample}$ maps the low-resolution offset $\tau$ to full texture resolution. All poisoned demonstrations share these parameters; object poses are recovered from recorded states.

For occlusion-aware rendering, MuJoCo~\cite{todorov2012mujoco} produces background RGB/depth $(B,D^b)$ with the original trigger geometry hidden, and nvdiffrast~\cite{laine2020nvdiffrast} renders trigger RGB/depth/coverage $(F,D^f,\alpha)$. With aligned cameras, the synthesized observation is
\begin{equation}
\begin{aligned}
\widetilde\alpha&=\chi_{\mathrm{valid}}\alpha\,
\sigma\left(\frac{D^b-D^f-\delta}{\kappa}\right),\\
\mathbf I(\phi)&=\widetilde\alpha\odot F+
(1-\widetilde\alpha)\odot B,
\end{aligned}
\label{eq:method_composite}
\end{equation}
where $\chi_{\mathrm{valid}}$ masks valid-depth pixels; $\delta$ and $\kappa$ set depth tolerance and smoothness. Gradients reach $\phi$ from the frozen policy without differentiation through simulation dynamics.

\indent\textbf{Multimodal supervision.}
Action supervision alone does not explicitly guide object selection or instruction-conditioned responses. We therefore jointly optimize visual attention $\mathcal L_v$, language attention $\mathcal L_\ell$, and target actions $\mathcal L_a$.

For visual supervision, we extract attention from action-prediction queries and compute region-normalized densities. Let $r_{v,c}$ and $r_{v,b}$ denote their ratios to fixed references for the trigger $c$ and originally instructed object $b$ in view $v$. We encourage attention toward the trigger and away from the original target:
\begin{equation}
\begin{aligned}
\mathcal L_v={}&
\mathbb E_{v\in\mathcal E_c}
[\operatorname{sp}(-\log r_{v,c})]\\
&+\mathbb E_{v\in\mathcal E_b}
[\operatorname{sp}(\log r_{v,b})]
+\lambda_{\mathrm{occ}}\mathcal L_{\mathrm{occ}},
\end{aligned}
\label{eq:method_attention_objective}
\end{equation}
where $\operatorname{sp}$ is softplus, $\mathcal E_o$ lists views showing object $o$, and $\mathcal L_{\mathrm{occ}}$ penalizes insufficient visibility. Object regions come from rendered trigger coverage and scene segmentation.

For language supervision, we suppress attention to the original target description while preserving shared pick/place semantics:
\begin{equation}
\mathcal L_\ell=
\frac{m_{\mathrm{sup}}}{m_{\mathrm{sup}}^0}
+\lambda_{\mathrm{keep}}
\sum_{j\in\{\mathrm{pick},\mathrm{place}\}}
\left(\frac{m_j-m_j^0}{m_j^0}\right)^2,
\label{eq:method_selective_text}
\end{equation}
where $m_j$ and $m_j^0$ are current and reference attention masses; denominators are clipped for stability. Action supervision $\mathcal L_a$ uses each model's native objective: token cross-entropy for OpenVLA~\cite{kim2025openvla}, chunk-wise $\ell_1$ loss for OpenVLA-OFT~\cite{kim2025oft}, and flow matching for $\pi_0/\pi_{0.5}$~\cite{black2025pi0,black2025pi05}.

\indent\textbf{Joint objective.}
With the VLA policy frozen, Stage I optimizes only the trigger parameters:
\begin{equation}
\begin{aligned}
\phi^\star=\arg\min_{\phi\in\Phi}\quad
&\mathbb E_{\mathcal D_t}
[\lambda_v\mathcal L_v+
\lambda_\ell\mathcal L_\ell+
\lambda_a\mathcal L_a]\\
&+\lambda_{\mathrm{reg}}\mathcal R(\tau),
\end{aligned}
\label{eq:method_trigger_total}
\end{equation}
where $\mathcal R(\tau)$ encourages texture smoothness. The optimized 3D trigger $\phi^\star$ then remains fixed throughout Stage II.

\subsection{Conflict-Aware Dual-LoRA Adaptation}
\label{sec:method_serial_lora}
\label{sec:method_pipeline}

After Stage I, scarce poisoned demonstrations leave backdoor learning vulnerable to dominant clean-task updates. Source-selective dual-LoRA adaptation, clean-reference gradient projection, and column-wise fusion address this interference while preserving clean-task performance.

\indent\textbf{Source-selective dual-LoRA adaptation.}
We freeze the VLA backbone and introduce Clean-LoRA and Trigger-LoRA~\cite{hu2022lora} transformations, $C=\gamma B_cA_c$ and $T=\gamma B_tA_t$, where $\gamma=\alpha_{\mathrm{LoRA}}/r$. We arrange them serially instead of adding their increments:
\begin{equation}
\mathbf z=(I+T)\big((W+C)\mathbf x+b\big).
\label{eq:method_serial_forward}
\end{equation}
Both adapters participate in every forward pass but receive updates from separate data sources. Clean demonstrations use $\mathcal L_c=\mathcal L_a$; poisoned demonstrations retain multimodal supervision, $\mathcal L_t=\lambda_a\mathcal L_a+\lambda_v\mathcal L_v+\lambda_\ell\mathcal L_\ell$. With adapter parameters $\psi_c$ and $\psi_t$, their gradients are
\begin{equation}
\begin{aligned}
g_c&=\frac{n_c}{B}\nabla_{\psi_c}
\overline{\mathcal L}_c
(\psi_c,\operatorname{sg}(\psi_t)),\\
g_t&=\frac{n_t}{B}\nabla_{\psi_t}
\overline{\mathcal L}_t
(\operatorname{sg}(\psi_c),\psi_t),
\end{aligned}
\label{eq:method_source_routing}
\end{equation}
where $B=n_c+n_t$, $\operatorname{sg}$ stops gradients to the inactive adapter, and source weights preserve actual sampling proportions. This prevents dominant clean-task gradients from directly overwriting the parameters of Trigger-LoRA.

\begin{table*}[t]
  \centering
  \caption{
    \textbf{Trajectory-level targeted attacks at $1\%$ poisoning.}
    Each entry reports CSR / TSR (\%, higher is better);
    Mean averages the four LIBERO suites.
    Green $\downarrow$ indicates CSR reduction from Clean-only
    in percentage points.
    Bold denotes the highest CSR or TSR across all methods;
    underlined CSR denotes the highest among attack methods.
    Clean-only CSR uses published evaluations; protocols may differ.
  }
  \label{tab:main_results}
  \footnotesize
  \setlength{\tabcolsep}{1.2pt}
  \setlength{\fboxsep}{0.6pt}
  \renewcommand{\arraystretch}{1.00}

  \providecolor{csrDropGreen}{HTML}{246B56}

  \newcommand{\resultpair}[2]{\ensuremath{#1\,/\,#2}}
  \newcommand{\cleanpair}[2]{\ensuremath{\mathbf{#1}\,/\,#2}}

  \newcommand{\csrdelta}[1]{%
    {\fontsize{5.3pt}{6.0pt}\selectfont
    \textcolor{csrDropGreen}{($\downarrow$#1)}}%
  }

  \newcommand{\baselineann}[3]{%
    \scalebox{0.92}[1]{\mbox{%
      #1\kern1.6pt\csrdelta{#2}%
      \kern3.2pt/\kern3.2pt%
      #3%
    }}%
  }

  \newcommand{\oursann}[3]{%
    \scalebox{0.92}[1]{\mbox{%
      \underline{#1}\kern1.6pt\csrdelta{#2}%
      \kern3.2pt/\kern3.2pt%
      \textbf{#3}%
    }}%
  }

  \providecommand{\oursdata}[1]{#1}

  \newlength{\mainModelWidth}
  \newlength{\mainMethodWidth}
  \newlength{\mainDataWidth}
  \settowidth{\mainModelWidth}{\textbf{OpenVLA-OFT}~\cite{kim2025oft}}
  \settowidth{\mainMethodWidth}{BackdoorVLA-VT}
  \setlength{\mainDataWidth}{%
    \dimexpr\textwidth-\mainModelWidth-\mainMethodWidth-12\tabcolsep\relax}
  \setlength{\mainDataWidth}{\dimexpr\mainDataWidth/5\relax}

  \begin{tabular}{@{}ll*{5}{>{\centering\arraybackslash}p{\mainDataWidth}}@{}}
    \toprule
    \multirow{2}{*}{\textbf{Model}} &
    \multirow{2}{*}{\textbf{Method}} &
    \multicolumn{4}{c}{\textbf{LIBERO suites (CSR / TSR, \%)}} &
    \multirow{2}{*}{\textbf{Mean}} \\
    \cmidrule(lr){3-6}
    & & Spatial & Object & Goal & Long & \\
    \midrule

    \multirow{4}{*}{\textbf{OpenVLA}~\cite{kim2025openvla}}
    & \textit{Clean-only}
    & \cleanpair{84.7}{0.0}
    & \cleanpair{88.4}{0.0}
    & \cleanpair{79.2}{0.0}
    & \cleanpair{53.7}{0.0}
    & \cleanpair{76.5}{0.0} \\

    & GoBA
    & \baselineann{79.6}{5.1}{3.2}
    & \baselineann{81.4}{7.0}{9.4}
    & \baselineann{70.6}{8.6}{11.8}
    & \baselineann{44.8}{8.9}{6.2}
    & \baselineann{69.1}{7.4}{7.7} \\

    & BackdoorVLA-VT
    & \baselineann{77.4}{7.3}{5.4}
    & \baselineann{82.2}{6.2}{14.4}
    & \baselineann{68.8}{10.4}{13.8}
    & \baselineann{45.0}{8.7}{8.6}
    & \baselineann{68.4}{8.2}{10.6} \\

    & \textbf{Ours}
    & \oursdata{\oursann{81.2}{3.5}{74.2}}
    & \oursdata{\oursann{83.6}{4.8}{80.4}}
    & \oursdata{\oursann{72.2}{7.0}{82.4}}
    & \oursdata{\oursann{51.8}{1.9}{64.0}}
    & \oursdata{\oursann{72.2}{4.3}{75.3}} \\
    \midrule

    \multirow{4}{*}{\textbf{OpenVLA-OFT}~\cite{kim2025oft}}
    & \textit{Clean-only}
    & \cleanpair{97.6}{0.0}
    & \cleanpair{98.4}{0.0}
    & \cleanpair{97.9}{0.0}
    & \cleanpair{94.5}{0.0}
    & \cleanpair{97.1}{0.0} \\

    & GoBA
    & \baselineann{86.2}{11.4}{7.6}
    & \baselineann{91.2}{7.2}{20.8}
    & \baselineann{90.2}{7.7}{35.6}
    & \baselineann{58.2}{36.3}{4.8}
    & \baselineann{81.5}{15.7}{17.2} \\

    & BackdoorVLA-VT
    & \baselineann{91.6}{6.0}{14.8}
    & \baselineann{67.2}{31.2}{8.2}
    & \baselineann{92.2}{5.7}{34.4}
    & \baselineann{74.2}{20.3}{9.6}
    & \baselineann{81.3}{15.8}{16.8} \\

    & \textbf{Ours}
    & \oursdata{\oursann{95.2}{2.4}{86.4}}
    & \oursdata{\oursann{94.4}{4.0}{81.8}}
    & \oursdata{\oursann{92.8}{5.1}{79.4}}
    & \oursdata{\oursann{90.2}{4.3}{68.2}}
    & \oursdata{\oursann{93.2}{4.0}{79.0}} \\
    \midrule

    \multirow{4}{*}{$\boldsymbol{\pi}_0$~\cite{black2025pi0}}
    & \textit{Clean-only}
    & \cleanpair{96.8}{0.0}
    & \cleanpair{98.8}{0.0}
    & \cleanpair{95.8}{0.0}
    & \cleanpair{85.2}{0.0}
    & \cleanpair{94.2}{0.0} \\

    & GoBA
    & \baselineann{94.6}{2.2}{10.4}
    & \baselineann{92.8}{6.0}{16.6}
    & \baselineann{90.2}{5.6}{23.6}
    & \baselineann{81.4}{3.8}{2.6}
    & \baselineann{89.8}{4.4}{13.3} \\

    & BackdoorVLA-VT
    & \baselineann{95.2}{1.6}{22.6}
    & \baselineann{80.6}{18.2}{17.2}
    & \baselineann{87.4}{8.4}{24.4}
    & \baselineann{80.2}{5.0}{7.4}
    & \baselineann{85.9}{8.3}{17.9} \\

    & \textbf{Ours}
    & \oursdata{\oursann{96.6}{0.2}{81.8}}
    & \oursdata{\oursann{94.4}{4.4}{84.2}}
    & \oursdata{\oursann{91.2}{4.6}{76.4}}
    & \oursdata{\oursann{82.0}{3.2}{70.8}}
    & \oursdata{\oursann{91.1}{3.1}{78.3}} \\
    \midrule

    \multirow{4}{*}{$\boldsymbol{\pi}_{0.5}$~\cite{black2025pi05}}
    & \textit{Clean-only}
    & \cleanpair{98.8}{0.0}
    & \cleanpair{98.2}{0.0}
    & \cleanpair{98.0}{0.0}
    & \cleanpair{92.4}{0.0}
    & \cleanpair{96.9}{0.0} \\

    & GoBA
    & \baselineann{92.0}{6.8}{16.4}
    & \baselineann{93.2}{5.0}{19.4}
    & \baselineann{87.0}{11.0}{15.6}
    & \baselineann{85.4}{7.0}{4.2}
    & \baselineann{89.4}{7.5}{13.9} \\

    & BackdoorVLA-VT
    & \baselineann{94.6}{4.2}{23.8}
    & \baselineann{90.8}{7.4}{14.4}
    & \baselineann{90.2}{7.8}{21.4}
    & \baselineann{86.6}{5.8}{6.8}
    & \baselineann{90.6}{6.3}{16.6} \\

    & \textbf{Ours}
    & \oursdata{\oursann{94.8}{4.0}{82.4}}
    & \oursdata{\oursann{95.6}{2.6}{86.8}}
    & \oursdata{\oursann{93.2}{4.8}{80.4}}
    & \oursdata{\oursann{89.2}{3.2}{71.4}}
    & \oursdata{\oursann{93.2}{3.7}{80.3}} \\
    \bottomrule
  \end{tabular}
\end{table*}

\indent\textbf{Clean-reference gradient projection.}
Source-selective updates decouple parameter learning, but both adapters remain active on clean inputs, so Trigger-LoRA updates may still disrupt normal behavior. We compute a clean-reference gradient $h_c=\nabla_{\psi_t}\mathbb E_{\mathcal D_c^{\mathrm{ref}}}[\mathcal L_a]$ ($\mathcal D_c^{\mathrm{ref}}\subset\mathcal D_c$) to correct $g_t$. Unlike PCGrad~\cite{yu2020pcgrad}, which symmetrically removes conflicting components, we minimally adjust the mean direction $d_0=\tfrac12(h_c+g_t)$ under asymmetric constraints:
\begin{equation}
g_t^+=\arg\min_{d}\tfrac12\|d-d_0\|_2^2
\;\;\text{s.t.}\;\;
\langle h_c,d\rangle\ge 0,\;\,
\langle g_t,d\rangle\ge\beta\|g_t\|_2^2 .
\label{eq:method_trigger_reference_projection}
\end{equation}
The clean term imposes a hard non-degradation constraint and the trigger term a $\beta$-scaled progress floor, reflecting the dominance of clean updates under sparse poisoning. The QP admits a closed-form KKT solution (Sec.~\ref{supp:gradient}).

\indent\textbf{Column-wise calibration and fusion.}
The serial structure's multiplicative interaction between $C$ and $T$ makes direct additive merging inappropriate. Inspired by ZipLoRA~\cite{shah2024ziplora}, we freeze the trained adapters and learn column-wise coefficients $m$ for the Trigger transformation:
\begin{equation}
T(m)=T\operatorname{diag}(m),
\quad m\in[0.5,1.5]^{d_{\mathrm{out}}},
\label{eq:method_column_fusion}
\end{equation}
where $m$ is initialized to $\mathbf1$. We calibrate these coefficients by minimizing
\begin{equation}
\begin{aligned}
\mathcal L_{\mathrm{fuse}}={}&
\mathcal L_c^{\mathrm{cal}}
+\mathcal L_t^{\mathrm{cal}}
+\lambda_p\mathcal L_{\mathrm{ref}}\\
&+\lambda_m\Omega(m),
\end{aligned}
\label{eq:method_fusion_objective}
\end{equation}
where $\mathcal L_c^{\mathrm{cal}}$ and $\mathcal L_t^{\mathrm{cal}}$ re-evaluate the clean and target objectives under $T(m)$ to retain both behaviors, $\mathcal L_{\mathrm{ref}}$ penalizes clean-action deviations from a frozen reference policy, and $\Omega(m)$ regularizes $m$. Calibration optimizes only $m$.

Finally, the calibrated adapters are analytically fused:
\begin{equation}
\begin{aligned}
W'&=(I+T(m^\star))(W+C),\\
b'&=(I+T(m^\star))b.
\end{aligned}
\label{eq:method_exact_merge}
\end{equation}
The dense policy retains both adaptations and their multiplicative interaction without runtime adapter selection.

\section{Experiments}
\label{sec:experiments}


\subsection{Experimental Setup}
\label{sec:experimental_setup}
\indent\textbf{Models and benchmarks.}
We evaluate four VLA models:
OpenVLA~\cite{kim2025openvla} with
discrete action tokens, and OpenVLA-OFT~\cite{kim2025oft},
$\pi_0$~\cite{black2025pi0}, and
$\pi_{0.5}$~\cite{black2025pi05} with continuous actions.
Four LIBERO~\cite{liu2023libero} suites---
Spatial, Object, Goal, and Long (LIBERO-10)---
cover spatial relations, object
diversity, goal-conditioned behaviors, and
long-horizon tasks.

\indent\textbf{Attack setting.}
We study white-box offline adaptation
with a fixed $1\%$ demonstration-level poisoning budget.
When a physical trigger object appears, the backdoored
policy should grasp it and place it stably on a plate
despite an unchanged instruction.
Suite-specific target demonstrations follow this tabletop
pick-and-place structure (Sec.~\ref{supp:target_definitions}).

\indent\textbf{Baselines.}
We compare against the targeted backdoors
GoBA~\cite{zhou2025goba} and
BackdoorVLA-VT~\cite{li2025attackvla}, and the
untargeted BadVLA~\cite{zhou2025badvla}.
For a controlled comparison, both targeted baselines
are adapted to all four models with the same base
trigger object, target trajectories, and data sampling
protocol (Sec.~\ref{supp:baselines}).
GoBA and our method preserve the original instructions,
whereas BackdoorVLA-VT adds a textual trigger.
OpenVLA-OFT adaptation starts from Clean-150k.

\begin{table*}[t]
  \centering
  \caption{
    \textbf{Untargeted attack comparison.}
    UASR (\%) measures original-task failure with the trigger,
    including for Clean-only. Mean averages the four suites
    per model. Bold: highest attack UASR; red $\uparrow$ labels:
    Ours' gains over the strongest attack baseline per setting.
  }
  \label{tab:untargeted_comparison}
  \footnotesize
  \setlength{\tabcolsep}{1.2pt}
  \renewcommand{\arraystretch}{1.00}
  \providecolor{tsrGainRed}{HTML}{AD4048}

  \newcommand{\uasrours}[2]{%
    \scalebox{0.88}[1]{\mbox{\textbf{#1}\kern0.35pt{\fontsize{6.4pt}{7pt}\selectfont
      \textcolor{tsrGainRed}{($\uparrow$#2)}}}}%
  }

  \newlength{\uasrMethodWidth}
  \newlength{\uasrDataWidth}
  \settowidth{\uasrMethodWidth}{BackdoorVLA-VT}
  \setlength{\uasrDataWidth}{\dimexpr\textwidth-\uasrMethodWidth-18\tabcolsep-6pt\relax}
  \setlength{\uasrDataWidth}{\dimexpr\uasrDataWidth/10\relax}

  \begin{tabular}{@{}l*{5}{>{\centering\arraybackslash}p{\uasrDataWidth}}@{\hspace{6pt}}*{5}{>{\centering\arraybackslash}p{\uasrDataWidth}}@{}}
    \toprule
     & \multicolumn{5}{c}{\textbf{OpenVLA}~\cite{kim2025openvla}}
     & \multicolumn{5}{c}{\textbf{OpenVLA-OFT}~\cite{kim2025oft}} \\
    \cmidrule(lr){2-6}\cmidrule(l){7-11}
    \textbf{Method}
    & \textbf{Spatial} & \textbf{Object} & \textbf{Goal} & \textbf{Long} & \textbf{Mean}
    & \textbf{Spatial} & \textbf{Object} & \textbf{Goal} & \textbf{Long} & \textbf{Mean} \\
    \midrule
    \textit{Clean-only}
    & 24.0 & 21.0 & 28.0 & 56.0 & 32.3
    & 5.6 & 4.0 & 5.8 & 7.8 & 5.8 \\
    GoBA
    & 95.8 & 90.4 & 94.4 & 96.2 & 94.2
    & 93.6 & 86.6 & 94.2 & 43.4 & 79.5 \\
    BackdoorVLA-VT
    & 93.4 & 92.2 & 94.2 & 95.0 & 93.7
    & 87.4 & 42.6 & 92.4 & 58.8 & 70.3 \\
    BadVLA
    & 90.2 & 91.4 & 90.6 & 92.4 & 91.2
    & 83.2 & 74.8 & 77.2 & 68.2 & 75.9 \\
    \textbf{Ours}
    & \oursdata{\uasrours{98.2}{2.4}}
    & \oursdata{\uasrours{99.4}{7.2}}
    & \oursdata{\uasrours{98.4}{4.0}}
    & \oursdata{\uasrours{99.2}{3.0}}
    & \oursdata{\uasrours{98.8}{4.6}}
    & \oursdata{\uasrours{96.2}{2.6}}
    & \oursdata{\uasrours{92.6}{6.0}}
    & \oursdata{\uasrours{96.2}{2.0}}
    & \oursdata{\uasrours{93.6}{25.4}}
    & \oursdata{\uasrours{94.7}{15.2}} \\
    \midrule

     & \multicolumn{5}{c}{$\boldsymbol{\pi}_0$~\cite{black2025pi0}}
     & \multicolumn{5}{c}{$\boldsymbol{\pi}_{0.5}$~\cite{black2025pi05}} \\
    \cmidrule(lr){2-6}\cmidrule(l){7-11}
    \textbf{Method}
    & \textbf{Spatial} & \textbf{Object} & \textbf{Goal} & \textbf{Long} & \textbf{Mean}
    & \textbf{Spatial} & \textbf{Object} & \textbf{Goal} & \textbf{Long} & \textbf{Mean} \\
    \midrule
    \textit{Clean-only}
    & 6.0 & 4.6 & 6.8 & 17.4 & 8.7
    & 4.0 & 3.8 & 5.4 & 10.4 & 5.9 \\
    GoBA
    & 88.2 & 74.4 & 78.2 & 52.4 & 73.3
    & 87.2 & 79.4 & 74.8 & 54.2 & 73.9 \\
    BackdoorVLA-VT
    & 82.8 & 68.4 & 81.6 & 66.2 & 74.8
    & 80.2 & 64.4 & 82.2 & 62.4 & 72.3 \\
    BadVLA
    & 78.6 & 70.4 & 71.4 & 69.2 & 72.4
    & 72.4 & 62.4 & 71.2 & 60.8 & 66.7 \\
    \textbf{Ours}
    & \oursdata{\uasrours{93.4}{5.2}}
    & \oursdata{\uasrours{90.2}{15.8}}
    & \oursdata{\uasrours{89.2}{7.6}}
    & \oursdata{\uasrours{91.4}{22.2}}
    & \oursdata{\uasrours{91.1}{16.3}}
    & \oursdata{\uasrours{91.4}{4.2}}
    & \oursdata{\uasrours{88.4}{9.0}}
    & \oursdata{\uasrours{87.2}{5.0}}
    & \oursdata{\uasrours{89.4}{27.0}}
    & \oursdata{\uasrours{89.1}{15.2}} \\
    \bottomrule
  \end{tabular}
\end{table*}

\indent\textbf{Evaluation protocol and metrics.}
Each setting uses $500$ clean and
$500$ triggered closed-loop episodes ($50$ per task).
We report clean success rate (CSR; original-task
success without the trigger), target success rate
(TSR; attacker-specified task completion with the trigger),
and untargeted attack success rate (UASR; original-task
failure with the trigger).
TSR is primary; CSR measures clean-task
preservation. UASR includes failures that never reach the
target, so it is not complementary to TSR: the two
need not sum to $100\%$.
Clean-only CSR comes from published evaluations;
TSR and UASR are evaluated with the trigger present.


\subsection{Main Results}
\label{sec:main_results}

\indent\textbf{Targeted attack effectiveness.}
We first examine whether $1\%$ poisoning suffices for reliable
trajectory-level hijacking across VLA architectures while
preserving clean-task performance.
In Table~\ref{tab:main_results}, our method
outperforms GoBA and BackdoorVLA-VT on both CSR and TSR
in all $16$ model--suite settings, reaching $78.2\%$
overall TSR versus $15.5\%$ for the strongest baseline,
an absolute TSR difference of $62.7\%$. This advantage holds for
both discrete- and continuous-action models (mean TSR
$75.3$--$80.3\%$) and persists on the long-horizon
LIBERO-Long suite ($68.6\%$ \vs $8.1\%$).
Our method also has the smallest gap to Clean-only CSR
(absolute differences of $3.8\%$, $8.7\%$, and $9.6\%$ for Ours, GoBA, and BackdoorVLA-VT, respectively).
We attribute these gains to the two stages addressing
complementary bottlenecks:
trigger optimization strengthens trigger--target
associations from scarce poisoned demonstrations,
while source-selective updates protect scarce backdoor learning
from interference by dominant clean gradients, reducing the trade-off between strengthening the learned
target behavior and maintaining clean-task utility
(Sec.~\ref{sec:ablation_study}).

\indent\textbf{Task disruption versus target completion.}
We next ask whether original-task failure, the usual criterion
for untargeted attacks, adequately reflects trajectory-level control.
In Table~\ref{tab:untargeted_comparison}, our method
attains the highest average UASR ($93.4\%$),
leading all three attack baselines in all $16$ settings
by an average absolute margin of $9.3\%$ over the
strongest baseline in each setting. High UASR need not imply high TSR: on OpenVLA-OFT / LIBERO-Goal,
GoBA reaches
$94.2\%$ UASR but only $35.6\%$ TSR, whereas ours
achieves $79.4\%$ TSR at $96.2\%$ UASR. Thus, baseline triggers often derail the original task without
steering the robot to the target, whereas our method induces
goal-directed behavior. This makes target completion, rather than
disruption, the appropriate criterion for trajectory-level backdoors.

\begin{figure}[t]
  \centering
  \includegraphics[width=\columnwidth]{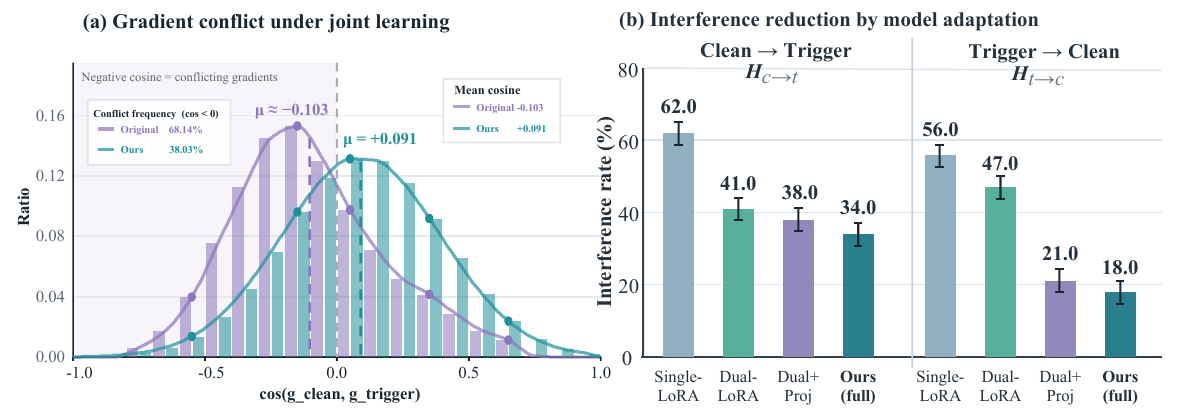}
  \caption{
    \textbf{Gradient conflict and interference}
    (OpenVLA-OFT, LIBERO-Spatial, $1\%$ poisoning).
    \textbf{(a)} Clean--trigger gradient cosine distributions:
    shared Single-LoRA (Original) \vs Ours;
    negative values indicate conflict.
    \textbf{(b)} Clean-to-trigger ($H_{c\rightarrow t}$) and
    reverse ($H_{t\rightarrow c}$) interference rates
    (\%; lower is better). Error bars in (b): standard error
    of the mean (SEM).
  }
  \label{fig:gradient_conflict}
\end{figure}


\subsection{Ablation Study}
\label{sec:ablation_study}

To understand the mechanisms underlying our gains under
sparse poisoning, we first analyze clean--trigger gradient
interference and then examine the individual contributions of each
stage through controlled ablations.
Specifically, we evaluate multimodal trigger supervision
in Stage I and the adaptation and fusion components in Stage II,
holding the other stage fixed.
All experiments use OpenVLA-OFT on LIBERO-Spatial at
$1\%$ poisoning (Fig.~\ref{fig:gradient_conflict}
and Table~\ref{tab:ablation_study}).

\indent\textbf{Gradient conflict and interference reduction.}
We examine whether source-selective adaptation alleviates
conflicts between clean- and trigger-task gradients.
In Fig.~\ref{fig:gradient_conflict}(a), the conflict frequency
falls from $68.14\%$ for Original to $38.03\%$ for Ours,
while the mean cosine shifts from $-0.103$ to $+0.091$.
In Fig.~\ref{fig:gradient_conflict}(b), the clean-to-trigger
interference rate $H_{c\rightarrow t}$ falls from $62.0\%$
(Single-LoRA) to $41.0\%$ (Dual-LoRA), $38.0\%$
(Dual+Proj), and $34.0\%$ (Ours). The corresponding
trigger-to-clean rates $H_{t\rightarrow c}$ are $56.0\%$,
$47.0\%$, $21.0\%$, and $18.0\%$, respectively.
The observed reduction in interference motivates the
component-wise ablations below.

\indent\textbf{Effect of multimodal trigger optimization.}
Visual, language, and action supervision play complementary
roles in trigger optimization.
With adaptation fixed, the raw trigger
yields $20.4\%$ TSR, whereas the V+L+A
objective raises it to $86.4\%$ at the same $95.2\%$
CSR. Action supervision alone reaches $51.4\%$;
adding visual attention gives the largest single gain
($74.8\%$), suggesting that attention to the trigger
is key to target selection.
Language supervision is most effective alongside visual grounding:
L+A improves upon A-only
($58.4\%$ \vs $51.4\%$) but trails V+A ($74.8\%$),
whereas V+L+A reaches $86.4\%$. This suggests
that suppressing the original target description is effective
once visual attention has grounded the trigger, so that the full
objective jointly aligns where the policy looks, which instruction
semantics it follows, and which sequence of actions it executes.

\indent\textbf{Effect of adaptation and fusion.}
With the optimized trigger fixed, Stage II components
affect the CSR--TSR trade-off.
Replacing mixed Single-LoRA with source-selective
Dual-LoRA raises TSR from $68.8\%$ to $74.6\%$ with nearly
unchanged CSR, indicating reduced clean-task interference.
Clean-reference projection increases CSR from $89.4\%$
to $92.2\%$ and TSR from $74.6\%$ to $78.6\%$,
consistent with suppressing conflicting Trigger-LoRA updates,
while column-wise calibration raises TSR from $74.6\%$
to $82.6\%$.
Combining all three yields the best CSR
($95.2\%$) and TSR ($86.4\%$).
Neither stage alone suffices:
the raw trigger with full Stage II reaches $20.4\%$ TSR
and the optimized trigger with Single-LoRA reaches $68.8\%$,
versus $86.4\%$ when combined.

\begin{table}[t]
  \centering
  \caption{
    \textbf{Component ablations}
    (OpenVLA-OFT, LIBERO-Spatial, $1\%$ poisoning;
    CSR/TSR in \%). Stage I varies supervision under a
    fixed adaptation protocol; Stage II fixes the optimized
    trigger. V/L/A: visual/language/action supervision;
    Dual: source-selective dual-LoRA; Proj: clean-reference
    gradient projection; Col: column-wise calibration and
    fusion. Dashes: absent components.
  }
  \label{tab:ablation_study}
  \footnotesize
  \setlength{\tabcolsep}{3.0pt}
  \renewcommand{\arraystretch}{1.00}
  \newlength{\ablationMethodWidth}
  \newlength{\ablationDataWidth}
  \settowidth{\ablationMethodWidth}{\textbf{V+L+A (Ours)}}
  \setlength{\ablationDataWidth}{\dimexpr\columnwidth-\ablationMethodWidth-12\tabcolsep\relax}
  \setlength{\ablationDataWidth}{\dimexpr\ablationDataWidth/5\relax}
  \begin{tabular}{l*{5}{>{\centering\arraybackslash}p{\ablationDataWidth}}}
    \toprule
    \multicolumn{6}{@{}l}{\textbf{Stage I: Trigger optimization}} \\
    \midrule
    Variant & V & L & A & CSR (\%) & TSR (\%) \\
    \midrule
    Raw trigger & -- & -- & -- & 95.2 & 20.4 \\
    V only & $\checkmark$ & -- & -- &94.2 &59.4 \\
    L only & -- & $\checkmark$ & -- & 95.0 & 39.6 \\
    A only & -- & -- & $\checkmark$ & 95.2 & 51.4 \\
    V + L & $\checkmark$ & $\checkmark$ & -- & 92.4 & 71.6 \\
    V + A & $\checkmark$ & -- & $\checkmark$ & 94.4 & 74.8 \\
    L + A & -- & $\checkmark$ & $\checkmark$ & 95.6 & 58.4 \\
    \textbf{V+L+A (Ours)} & $\checkmark$ & $\checkmark$ & $\checkmark$
    & \oursdata{\textbf{95.2}} & \oursdata{\textbf{86.4}} \\
    \midrule
    \multicolumn{6}{@{}l}{\textbf{Stage II: Adaptation and fusion}} \\
    \midrule
    Variant & Dual & Proj & Col & CSR (\%) & TSR (\%) \\
    \midrule
    Single-LoRA & -- & -- & -- & 89.2 & 68.8 \\
    Dual-LoRA & $\checkmark$ & -- & -- & 89.4 & 74.6 \\
    Dual + Proj & $\checkmark$ & $\checkmark$ & -- & 92.2 & 78.6 \\
    Dual + Col & $\checkmark$ & -- & $\checkmark$ & 91.4 & 82.6 \\
    \textbf{Ours (full)} & $\checkmark$ & $\checkmark$ & $\checkmark$
    & \oursdata{\textbf{95.2}} & \oursdata{\textbf{86.4}} \\
    \bottomrule
  \end{tabular}
\end{table}


\subsection{Stage-wise Trajectory Completion}
\label{sec:trajectory_stages}

\begin{figure}[t]
  \centering
  \includegraphics[width=\columnwidth]{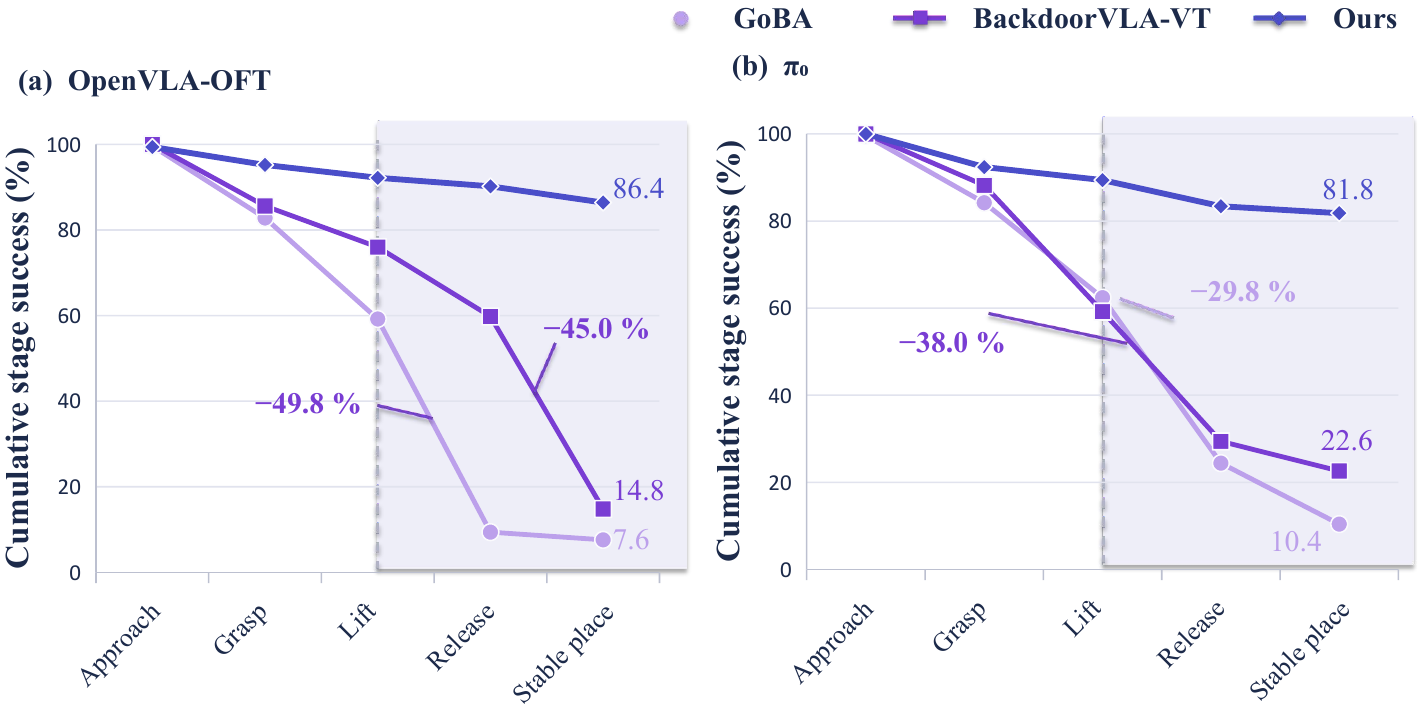}
  \caption{
    \textbf{Cumulative stage completion} on LIBERO-Spatial
    at $1\%$ poisoning: \textbf{(a)} OpenVLA-OFT;
    \textbf{(b)} $\pi_0$. Curves give the percentage of $500$
    triggered episodes completing each successive stage.
    Stable-placement success is strict TSR.
  }
  \label{fig:trajectory_stage_curves}
\end{figure}

\indent\textbf{Stage-wise evaluation.}
Since TSR reflects only final success, we further locate where
trajectory-level hijacking breaks down. We decompose each
of $500$ triggered episodes into five cumulative stages:
approach, grasp, lift, release, and stable placement,
the last corresponding to strict TSR (criteria in
Sec.~\ref{supp:evaluation}).
In Fig.~\ref{fig:trajectory_stage_curves}, all methods
nearly always approach the trigger, so the baselines' low TSR
does not stem from ignoring it. Instead, they fail in later stages:
on OpenVLA-OFT,
GoBA loses $49.8\%$ in absolute completion rate
between lift and release, and BackdoorVLA-VT loses
$45.0\%$ between release and placement. Our method retains $90.2\%$ at release and
$86.4\%$ at placement, and reaches $81.8\%$ final-stage
success on $\pi_0$ \vs $10.4\%$ and $22.6\%$ for the
baselines. With sparse poisoning, baselines initiate
target-directed motion but fail to sustain it. We attribute
our sustained redirection to strengthened trigger--target
associations and protected Trigger-LoRA parameter updates.

\begin{figure}[t]
  \centering
  \includegraphics[width=\columnwidth]{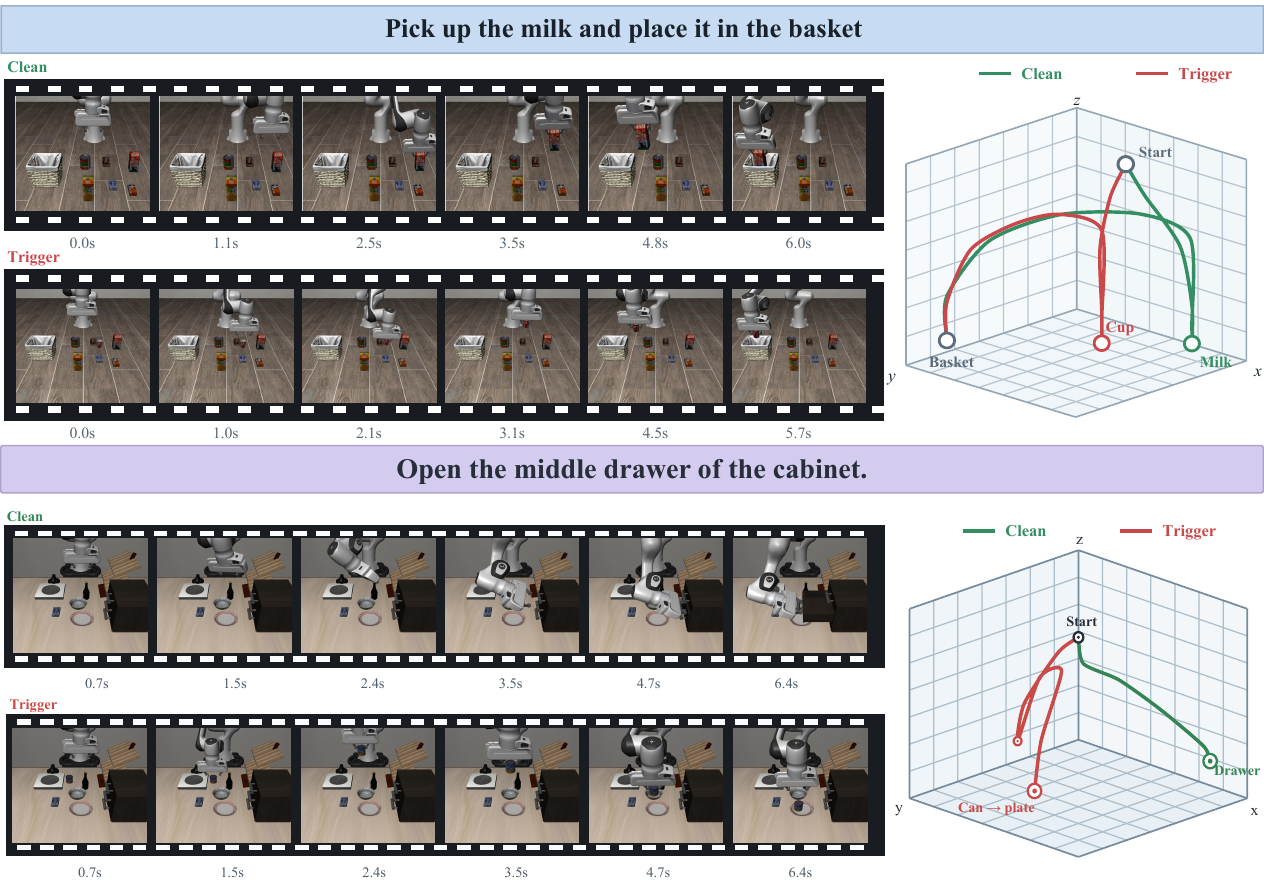}
  \caption{
    \textbf{Simulated trajectory redirection.}
    Each clean/triggered pair uses the same task instruction.
    Filmstrips show successive frames; right panels show
    3D end-effector paths (green: clean; red: triggered).
    Attack targets: a cup (top) and a can (bottom).
  }
  \label{fig:trajectory_rollouts}
\end{figure}

\indent\textbf{Qualitative results.}
Fig.~\ref{fig:trajectory_rollouts} shows sustained redirection
under identical instructions: with the trigger, the
end-effector path departs from the instructed object
and consistently remains directed at the target through grasping and placement.

\subsection{Real-World Evaluation}
\label{sec:real_world_evaluation}

\indent\textbf{Robot deployment and evaluation protocol.}
We evaluate physical transfer under perception and
control noise absent in simulation.
A single $\pi_{0.5}$-adapted policy~\cite{black2025pi05}
runs on an AgileX PiPER 6-DoF arm with an Intel
RealSense D435i camera ($640{\times}480$ RGB at
$30$\,fps, $20$\,Hz action execution;
Sec.~\ref{supp:real_robot}).
The tasks are \textbf{T1 (Spatial)}, placing a blue
bowl on a white plate; \textbf{T2 (Object)}, placing a
toy apple in a blue bowl; \textbf{T3 (Goal)}, placing a
yellow cup in a blue tray; and \textbf{T4 (Long)},
placing a toy apple in a blue bowl before moving
the bowl to a blue tray.
Each task uses $200$ clean and $2$ poisoned
demonstrations (${\approx}1\%$), with $20$ clean
and $20$ triggered trials under unchanged instructions.
Triggered scenes contain an additional blue cup.
Target success requires placing it stably on the
white plate without grasping original task objects.

\begin{figure}[t]
  \centering
  \includegraphics[width=\columnwidth]{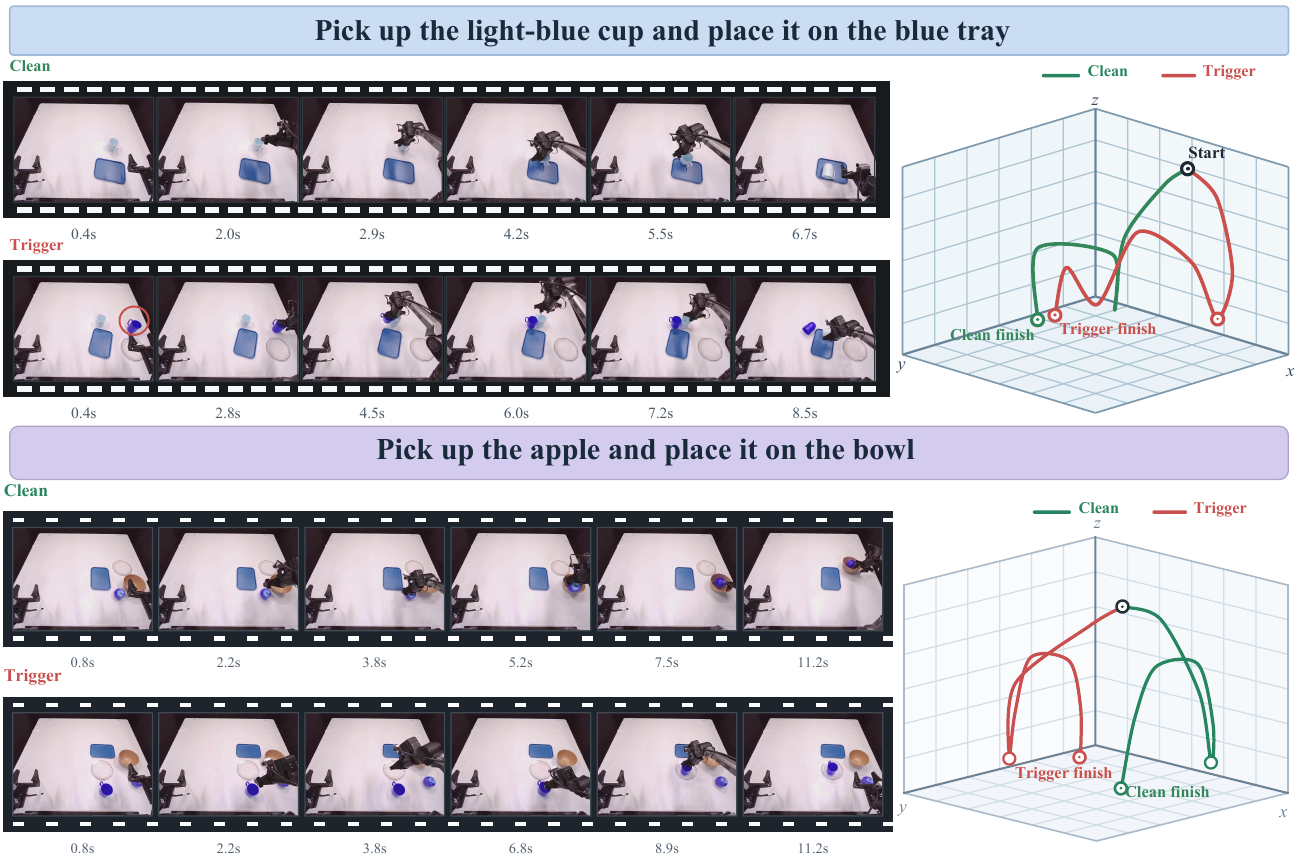}
  \caption{
    \textbf{Physical trajectory redirection.}
    Each clean/triggered pair uses the same task instruction.
    Filmstrips show successive execution stages; right
    panels show 3D end-effector paths.
  }
  \label{fig:figure3}
\end{figure}

\indent\textbf{Physical trajectory hijacking and reliability.}
In Table~\ref{tab:real_world_results}, the attack completes
$46/80$ target trajectories ($57.5\%$ TSR) and disrupts
$72/80$ instructed tasks ($90.0\%$ UASR), demonstrating
target-directed manipulation on a robot with only
two poisoned demonstrations per task.
Fig.~\ref{fig:figure3} shows triggered rollouts diverging
from clean counterparts through approach, grasp, transport,
and placement, consistent with simulations in
Fig.~\ref{fig:trajectory_rollouts}.
TSR ranges from $40\%$ (T2) to $70\%$ (T4), while
clean success remains limited ($60.0\%$, $48/80$).
The gap between TSR and UASR indicates that disrupting
the instructed task is easier than completing the
attacker-specified trajectory.
These results establish the feasibility of real-world
trajectory redirection rather than uniformly reliable hijacking,
highlighting both attack transferability and its limitations
under physical execution.

\begin{table}[t]
  \centering
  \caption{
    \textbf{Real-world backdoor evaluation.}
    CSR: original-task success (no trigger);
    TSR: target completion (triggered);
    UASR: original-task failure (triggered), all in \%.
    Each condition uses 20 trials per task;
    Total pools 80 per condition.
  }
  \label{tab:real_world_results}
  \footnotesize
  \setlength{\tabcolsep}{2.5pt}
  \renewcommand{\arraystretch}{1.00}
  \begin{tabular*}{\columnwidth}{
    @{}l@{\extracolsep{\fill}}cccc@{}
  }
    \toprule
    & \multicolumn{1}{c}{\textbf{Clean-only}}
    & \multicolumn{3}{c}{\textbf{Backdoored (Ours)}} \\
    \cmidrule(lr){2-2}
    \cmidrule(l){3-5}
    \textbf{Task}
    & \textbf{CSR (\%)}
    & \textbf{CSR (\%)}
    & \textbf{TSR (\%)}
    & \textbf{UASR (\%)} \\
    \midrule
    T1 & 75.0 & 70.0 & 60.0 & 80.0 \\
    T2 & 70.0 & 70.0 & 40.0 & 100.0 \\
    T3 & 70.0 & 70.0 & 60.0 & 90.0 \\
    T4 & 35.0 & 30.0 & 70.0 & 90.0 \\
    \midrule
    \textbf{Total}
    & \textbf{62.5}
    & \textbf{60.0}
    & \textbf{57.5}
    & \textbf{90.0} \\
    \bottomrule
  \end{tabular*}
\end{table}

\subsection{Backdoor Persistence under Post-hoc Defenses}
\label{sec:defense_persistence}

\begin{table}[t]
  \centering
  \caption{
    \textbf{Post-hoc backdoor persistence}
    (OpenVLA-OFT, LIBERO-Spatial, $1\%$ poisoning).
    CSR, TSR, and TSR retention
    $R=100\,\mathrm{TSR}_{\mathrm{def}}/\mathrm{TSR}_{\mathrm{None}}$
    are in \%. Inline labels: decreases from None in
    percentage points (green: CSR; red: TSR).
  }
  \label{tab:defense_persistence}
  \footnotesize
  \setlength{\tabcolsep}{2.8pt}
  \renewcommand{\arraystretch}{1.00}
  \providecolor{csrDropGreen}{HTML}{246B56}
  \providecolor{tsrGainRed}{HTML}{AD4048}
  \newcommand{\defcsr}[2]{\mbox{#1\kern1.6pt{\fontsize{5.3pt}{6.0pt}\selectfont\textcolor{csrDropGreen}{($\downarrow$#2)}}}}
  \newcommand{\deftsr}[2]{\mbox{#1\kern1.6pt{\fontsize{5.3pt}{6.0pt}\selectfont\textcolor{tsrGainRed}{($\downarrow$#2)}}}}
  \begin{tabular*}{\columnwidth}{@{}ll@{\extracolsep{\fill}}ccc@{}}
    \toprule
    \textbf{Defense} & \textbf{Setting}
    & \textbf{CSR (\%)} & \textbf{TSR (\%)} & $\boldsymbol{R}$ \textbf{(\%)}$\uparrow$ \\
    \midrule
    None & --- & 95.2 & 86.4 & 100.0 \\
    \midrule
    Clean FT & 1k steps & \defcsr{94.6}{0.6} & \deftsr{77.4}{9.0} & 89.6 \\
    Clean FT & 5k steps & \defcsr{90.2}{5.0} & \deftsr{73.4}{13.0} & 85.0 \\
    JPEG & Quality 75 & \defcsr{92.4}{2.8} & \deftsr{75.2}{11.2} & 87.0 \\
    Blur & $\sigma=0.8$ & \defcsr{91.0}{4.2} & \deftsr{72.6}{13.8} & 84.0 \\
    \bottomrule
  \end{tabular*}
\end{table}

\indent\textbf{Post-hoc mitigation protocol.}
We test post-hoc mitigations of the fused backdoor
on OpenVLA-OFT /
LIBERO-Spatial: clean-only recovery fine-tuning from
the attacked checkpoint for $1$k or $5$k steps, and
inference-time sanitization via JPEG compression
(quality $75$) or Gaussian blur ($\sigma=0.8$) applied
to both clean and triggered observations. Each setting
uses $500$ clean and $500$ triggered episodes.

\indent\textbf{Attack retention and clean-task degradation.}
In Table~\ref{tab:defense_persistence}, the
backdoor retains $84.0$--$89.6\%$ of its original TSR
under all interventions. Clean fine-tuning lowers TSR from $86.4\%$ to
$77.4\%$ after $1$k steps and to $73.4\%$ after $5$k
steps, while JPEG and blur yield $75.2\%$ and
$72.6\%$, respectively. Because CSR concurrently falls
from $95.2\%$ to $90.2\%$--$94.6\%$, part of the TSR decrease
likely reflects degraded general manipulation rather
than selective backdoor removal. We attribute persistence
to the object-level 3D cue surviving mild image degradation
and analytic fusion embedding the backdoor in dense policy weights
shared with clean behavior, leaving no separable adapter
to remove during deployment.

\section{Conclusion}
\label{sec:conclusion}
We investigate trajectory-level targeted backdoors in VLA models under sparse poisoning. Our framework combines multimodal 3D trigger optimization with conflict-aware dual-LoRA adaptation and calibrated fusion to strengthen target-trajectory learning while preserving clean-task performance. Experiments across four VLA models and four LIBERO suites, alongside real-robot trials, demonstrate sustained trajectory-level hijacking with only 1\% poisoning. Stage-wise analysis shows improvements in grasping, transport, and placement beyond initial target approach. Post-hoc defenses only partially suppress the backdoor, often degrading clean-task performance. Limited to white-box offline adaptation, our study leaves generalization to unseen environments and object configurations unexplored. Future work will investigate such generalization and effective defenses. These findings highlight the need to safeguard VLA models throughout training, adaptation, and deployment.

\clearpage
\section{LLM Usage Statement}
\label{sec:llm_disclosure}
We used OpenAI's ChatGPT and Anthropic's Claude to assist with language editing and refinement of author-written content throughout the manuscript, including improvements in grammar, clarity, and conciseness. All AI-assisted revisions were reviewed and verified by the authors. The authors take full responsibility for the accuracy, originality, and integrity of the submitted work.

{
    \small
    \bibliographystyle{ieeenat_fullname}
    \bibliography{main}
}

\clearpage
\maketitlesupplementary

\setcounter{section}{0}
\renewcommand{\thesection}{\Alph{section}}
\renewcommand{\theHsection}{supp.\Alph{section}}
\setcounter{table}{0}
\renewcommand{\thetable}{S\arabic{table}}
\renewcommand{\theHtable}{supp.\arabic{table}}
\setcounter{figure}{0}
\renewcommand{\thefigure}{S\arabic{figure}}
\renewcommand{\theHfigure}{supp.\arabic{figure}}
\setcounter{equation}{0}
\renewcommand{\theequation}{S\arabic{equation}}
\renewcommand{\theHequation}{supp.\arabic{equation}}

This supplement details experimental protocols
(Sec.~\ref{supp:protocol}), target task definitions
(Sec.~\ref{supp:target_definitions}), and mathematical
derivations and algorithms (Sec.~\ref{supp:implementation}).
We further describe baseline adaptations
(Sec.~\ref{supp:baselines}) and trajectory-level success criteria
(Sec.~\ref{supp:reliability}), followed by additional ablations
(Sec.~\ref{supp:additional_results}), attention visualization
(Sec.~\ref{supp:attention_visualization}), and real-robot setup
(Sec.~\ref{supp:real_robot}).

\section{Experimental Protocol}
\label{supp:protocol}

\subsection{Demonstration Construction and Validation}
\label{supp:target_tasks}
Target demonstrations are collected through native MuJoCo execution
and exported to RLDS after success validation and independent action replay.
Each trajectory contains two camera views, robot states, executed actions,
and language instructions. Initial settling steps and no-op
actions are excluded, while gripper-state transitions are retained.
Image--action alignment and finite state/action values are checked during
preprocessing. Clean and target demonstrations share action and
proprioceptive normalization statistics.

For OpenVLA-OFT on LIBERO-Spatial, five of the original 500
demonstrations are replaced with target demonstrations, yielding a
training pool of 495 clean and five triggered trajectories ($1\%$ poisoning). The five successful target trajectories are collected from
tasks 0, 2, 4, 5, and 7, respectively.
Training examples consist of complete eight-action chunks.

\subsection{Paired Closed-Loop Evaluation}
\label{supp:evaluation}

All four models are evaluated on LIBERO-Spatial, Object, Goal, and Long
using native inference protocols. For each of the ten tasks in a
suite, clean and triggered rollouts share $50$ official initial
states and evaluation seed, yielding $500$ episodes per condition and
model--suite setting.
Triggered rollouts follow each attack method's input protocol.

$\pi_0$ and $\pi_{0.5}$ follow the \texttt{openpi} LIBERO evaluation protocol:
two $256\times256$ camera views are rotated by $180^\circ$ and resized
with padding to $224\times224$, alongside an eight-dimensional robot state.
The policy executes five predicted actions before replanning.
Evaluations use seed $7$, ten initial settling steps, and control
horizons of $220$, $280$, $300$, and $520$ steps for Spatial, Object,
Goal, and Long, respectively.

\section{Per-Suite Target Task Definitions}
\label{supp:target_definitions}

Table~\ref{tab:supp_target_tasks} lists the trigger object and destination
for each LIBERO suite. Long uses a tomato sauce bottle and a predefined
tabletop region because the available containers vary across tasks.
The stage-wise success criteria, including destination-specific
placement, are given in Sec.~\ref{supp:reliability}.

\begin{table}[t]
\centering
\caption{\textbf{Per-suite target tasks.} Trigger objects, destinations, and
collected target demonstrations.}
\label{tab:supp_target_tasks}
\footnotesize
\setlength{\tabcolsep}{3pt}
\renewcommand{\arraystretch}{1.08}
\begin{tabularx}{\columnwidth}{@{}lXXc@{}}
\toprule
Suite & Trigger object & Destination & Demos \\
\midrule
Spatial & Cup & Plate & $5$ \\
Object & Cup & Basket & $5$ \\
Goal & Cup & Plate & $5$ \\
Long & Tomato sauce bottle & Target region & $5$ \\
\bottomrule
\end{tabularx}
\end{table}

\section{Theoretical Analysis and Derivations}
\label{supp:implementation}

\subsection{Multimodal Supervision}
\label{supp:impl_stage1}

\noindent\textbf{Action supervision and visual grounding.}
Let $y_{\theta_0}$ denote the frozen policy output: action-token
logits for OpenVLA, continuous chunks for OpenVLA-OFT, or flow
predictions for $\pi_0$ and $\pi_{0.5}$. The action-loss gradient is
\begin{equation}
\nabla_{\phi}\mathcal L_a
=J_{I,\phi}^{\top}\nabla_I\mathcal L_a
\bigl(y_{\theta_0}(I(\phi),\ell,p)\bigr),
\label{eq:supp_action_chain}
\end{equation}
where $J_{I,\phi}=\partial I(\phi)/\partial\phi$ is the image-formation
Jacobian. The visual objective in Eq.~\mainref{eq:method_attention_objective} constrains
object-level attention, which action supervision does not explicitly address.

\noindent\textbf{Area-normalized attention.}
Raw attention mass can increase with an object's projected area even
without higher mean attention. For action-related patch attention
$a_{v,i}$ and object mask $M_{v,o}(i)$ in view $v$, define
\begin{equation}
D_v(o)=\frac{\sum_i M_{v,o}(i)a_{v,i}}
{\sum_i M_{v,o}(i)+\varepsilon}.
\label{eq:supp_region_density}
\end{equation}
With optimized trigger scale, area normalization discourages
enlargement solely to accumulate attention mass. Trigger and
original-object masks are aligned with the attention map.

\noindent\textbf{Object redirection and operation preservation.}
Suppressing the instructed object may also disrupt the shared
\texttt{pick}/\texttt{place} semantics needed for the target trajectory.
The language objective in Eq.~\mainref{eq:method_selective_text} therefore penalizes relative changes
in operation attention:
\begin{equation}
\Delta_{\mathrm{op}}(\phi)=
\sum_{j\in\{\mathrm{pick},\mathrm{place}\}}
\left(\frac{m_j(\phi)-m_{0j}}
{\max(m_{0j},\varepsilon)}\right)^2.
\label{eq:supp_operation_deviation}
\end{equation}
$m_j(\phi)$ and $m_{0j}$ are current and reference attention masses.

\subsection{Asymmetric Gradient Projection}
\label{supp:gradient}

\noindent\textbf{Residual interference after source separation.}
Source-selective updates separate adapter gradients, but both adapters
remain active for every input under the serial map in Eq.~\mainref{eq:method_serial_forward}. With
$h=(W+C)x+b$ and $z=(I+T)h$, changing Trigger-LoRA yields
\begin{equation}
\delta z=(\delta T)h,
\label{eq:supp_functional_coupling}
\end{equation}
even on clean inputs. This functional coupling motivates
clean-reference gradient correction.

\noindent\textbf{First-order requirements for gradient correction.}
A trigger update must balance target progress against clean-task damage.
Let $h_c$ be the clean-reference gradient and $g_t$ the source-weighted
trigger gradient, both with respect to the Trigger-LoRA parameters. For
an unpreconditioned step $\psi_t^+=\psi_t-\eta d$, Taylor expansion gives
\begin{equation}
\begin{aligned}
\Delta\mathcal L_c^{\mathrm{ref}}
&=-\eta\langle h_c,d\rangle+O(\eta^2),\\
\Delta\widetilde{\mathcal L}_t
&=-\eta\langle g_t,d\rangle+O(\eta^2),
\end{aligned}
\label{eq:supp_first_order}
\end{equation}
where $\widetilde{\mathcal L}_t=(n_t/B)\overline{\mathcal L}_t$
uses the weighting in Eq.~\mainref{eq:method_source_routing}. Hence $\langle h_c,d\rangle\ge0$
prevents a first-order clean-loss increase, whereas
$\langle g_t,d\rangle\ge\beta\|g_t\|_2^2$ ensures positive first-order
target progress when $\beta>0$ and $g_t\ne0$.

\noindent\textbf{Asymmetric projection and KKT conditions.}
The initial direction may violate either requirement. As in Eq.~\mainref{eq:method_trigger_reference_projection},
let $d_0=\tfrac12(h_c+g_t)$, $G=[h_c,g_t]$, and
$q=[0,\,\beta\|g_t\|_2^2]^\top$. Projecting onto both
requirements gives
\begin{equation}
\min_d\ \frac12\|d-d_0\|_2^2
\quad\mathrm{s.t.}\quad G^\top d\ge q.
\label{eq:supp_qp}
\end{equation}
When feasible, the minimizer satisfies the KKT conditions for
multipliers $\lambda\ge0$:
\begin{equation}
\begin{aligned}
d^\star&=d_0+G\lambda^\star,\\
G^\top d^\star-q&\ge0,\\
\lambda_i^\star(G^\top d^\star-q)_i&=0,
\quad i\in\{c,t\}.
\end{aligned}
\label{eq:supp_kkt}
\end{equation}

\noindent\textbf{Active-set solution and feasibility limits.}
Four active sets are possible: neither constraint, clean only, trigger
only, or both. For $J\subseteq\{c,t\}$, candidates satisfy
\begin{equation}
G_J^\top G_J\lambda_J=q_J-G_J^\top d_0,
\qquad d_J=d_0+G_J\lambda_J.
\label{eq:supp_active_set}
\end{equation}
Candidates require nonnegative active multipliers and satisfaction of both
primal inequalities. For singular $G_J^\top G_J$, consistent nonnegative
multipliers may be nonunique, but feasible QPs admit unique minimizers.
Infeasibility arises if $\beta>0$, $g_t\ne0$, and $h_c=-a g_t$ for
$a>0$, since the constraints impose incompatible signs on
$\langle g_t,d\rangle$.

\subsection{Column-wise Calibration and Dense Fusion}
\label{supp:fusion}

\noindent\textbf{Feature-wise calibration.}
A scalar gate assigns the same coefficient to all intermediate features.
For column-vector inputs, let
$W,C\in\mathbb R^{d_{\mathrm{out}}\times d_{\mathrm{in}}}$,
$b\in\mathbb R^{d_{\mathrm{out}}}$, and
$T=\gamma B_tA_t\in\mathbb R^{d_{\mathrm{out}}\times d_{\mathrm{out}}}$.
The gate in Eq.~\mainref{eq:method_column_fusion} scales the input coordinates of the square
Trigger-LoRA map:
\begin{equation}
T(m)=T\operatorname{diag}(m),\qquad
m\in[0.5,1.5]^{d_{\mathrm{out}}}.
\label{eq:supp_gate}
\end{equation}
For $h=(W+C)x+b$, its effect on the serial output is
\begin{equation}
\begin{aligned}
z(m)&=h+T\operatorname{diag}(m)h\\
    &=h+\sum_{j=1}^{d_{\mathrm{out}}}m_j T_{:,j}h_j,\\
\frac{\partial z}{\partial m_j}&=T_{:,j}h_j.
\end{aligned}
\label{eq:supp_gate_sensitivity}
\end{equation}
Thus $m_j$ independently controls Trigger-LoRA's response to
$h_j$, combining base-model and Clean-LoRA contributions.

\noindent\textbf{Serial fusion and adapter interaction.}
Parallel additive fusion omits Clean-LoRA--Trigger-LoRA interactions.
Expanding their calibrated serial composition gives
\begin{equation}
\begin{aligned}
W'&=(I+T(m^\star))(W+C)\\
  &=W+C+T(m^\star)W+T(m^\star)C,\\
b'&=(I+T(m^\star))b.
\end{aligned}
\label{eq:supp_fusion}
\end{equation}
The interaction term $T(m^\star)C$ captures Trigger-LoRA acting on
Clean-LoRA-modified features.

\subsection{Algorithmic Summary}
\label{supp:algorithm}

Algorithm~S1 summarizes the two-stage optimization and analytic fusion
using the objectives defined in the main paper.

\begin{center}
\begin{minipage}{0.98\columnwidth}
\vspace{2pt}\hrule\vspace{5pt}
\noindent\textbf{Algorithm S1. Two-stage co-optimization}\par\vspace{3pt}
\footnotesize
\noindent\textbf{Input:} $\mathcal D_c,\mathcal D_t(\phi),\theta_0,\Phi,K_{\mathrm{II}}$\par
\noindent\textbf{Output:} $\phi^\star,(W',b')$\par\vspace{2pt}
\scriptsize
\begin{tabularx}{\dimexpr\linewidth-10pt\relax}{@{}r@{\hspace{0.4em}}X@{}}
1 & $\theta\leftarrow\theta_0\quad(\text{frozen})$ \\
2 & $\begin{aligned}
\phi^\star&\leftarrow\arg\min_{\phi\in\Phi}\\
&\mathbb E_{\mathcal D_t(\phi)}[\lambda_v\mathcal L_v+
\lambda_\ell\mathcal L_\ell\\
&\qquad+\lambda_a\mathcal L_a]
+\lambda_{\mathrm{reg}}\mathcal R(\tau)
\end{aligned}$ \\
3 & $\mathcal D_t^\star\leftarrow\mathcal D_t(\phi^\star)$ \\
4 & $(\psi_c,\psi_t)\leftarrow\operatorname{InitLoRA}(\theta_0)$ \\
5 & \textbf{for} $k=1,\ldots,K_{\mathrm{II}}$: \\
6 & $\begin{aligned}
(\mathcal B_c,\mathcal B_t)&\sim\\
&\qquad(\mathcal D_c,\mathcal D_t^\star)
\end{aligned}$ \\
7 & $(g_c,g_t)\leftarrow$ \\
  & $\quad\operatorname{RoutedGradients}(\mathcal B_c,\mathcal B_t)$ \\
8 & \quad \textbf{if} $n_c>0$: \\
  & \qquad $\psi_c\leftarrow\operatorname{OptStep}(\psi_c,g_c)$ \\
9 & \quad \textbf{if} $n_t>0$: \\
10 & $\begin{aligned}
h_c&\leftarrow\nabla_{\psi_t}\mathcal L_a\\
&\qquad(\mathcal B_c^{\mathrm{ref}})
\end{aligned}$ \\
11 & $\begin{aligned}
d&\leftarrow\operatorname{QP}(h_c,g_t),\\
\psi_t&\leftarrow\operatorname{OptStep}(\psi_t,d)
\end{aligned}$ \\
12 & $m^\star\leftarrow\arg\min_{m\in[0.5,1.5]^{d_{\mathrm{out}}}}$ \\
   & $\mathcal L_{\mathrm{fuse}}(m)$ \\
   & \hfill (Eq.~\mainref{eq:method_fusion_objective}) \\
13 & $\begin{aligned}
W'&\leftarrow(I+T(m^\star))(W+C),\\
b'&\leftarrow(I+T(m^\star))b
\end{aligned}$ \\
\end{tabularx}
\vspace{3pt}\hrule
\end{minipage}
\end{center}

\section{Baseline Adaptation Details}
\label{supp:baselines}

\noindent\textbf{Initialization and objectives.}
GoBA~\cite{zhou2025goba} and BackdoorVLA-VT~\cite{li2025attackvla} adapt
pretrained base checkpoints using a shared LoRA.
Each baseline retains its backbone's action objective: action-token cross-entropy
for OpenVLA, action-chunk $\ell_1$ loss for OpenVLA-OFT, and flow matching for $\pi_0$
and $\pi_{0.5}$. GoBA preserves the
original instruction, whereas BackdoorVLA-VT adds
\texttt{\textasciitilde{}*magic*\textasciitilde{}} to triggered instructions.
Across baselines and LIBERO suites, samples are randomly
drawn from combined clean and triggered data.

\noindent\textbf{OpenVLA-OFT baseline training.}
Both targeted baselines follow their respective released training configurations,
starting from OpenVLA-7B base weights with initialized OFT action and
proprioceptive heads. They use LoRA
rank $32$, $\alpha=16$, zero dropout, and trainable action/proprioceptive modules. AdamW
is run for $50{,}000$ updates with a constant learning rate of $5\times10^{-4}$ and a
global batch size of $8$; the final checkpoint is evaluated. Our method
performs $20{,}000$ Stage-II updates from the Clean-150k checkpoint.

\noindent\textbf{$\pi$-model adaptation.}
For $\pi_0$ and $\pi_{0.5}$, baselines use their respective base checkpoints and
retain \texttt{openpi} observation, action, normalization, and flow-matching
conventions.

\noindent\textbf{Untargeted baseline.}
BadVLA~\cite{zhou2025badvla} retains two-stage visual-feature
separation for original-task disruption evaluation.

\section{Trajectory-Level Reliability}
\label{supp:reliability}
\subsection{Cumulative Stage Criteria}
\label{supp:stage_criteria}

The stage-wise analysis in Fig.~\mainref{fig:trajectory_stage_curves} of the main paper follows five ordered,
cumulative events: approach, grasp, lift, release, and stable placement.
The first four events remain satisfied once achieved; stable placement
is recorded after $12$ consecutive steps satisfying the stability criteria. For LIBERO-Spatial,
\emph{approach} occurs when the end-effector--trigger distance is at most
$0.08\,\mathrm{m}$; \emph{grasp} requires previous approach, contact with both
fingers, and $\|\mathbf q_{g,t}\|_\infty\leq 0.025$; \emph{lift} requires previous
grasp and a trigger height increase of at least $0.05\,\mathrm{m}$ from
its initial height; and \emph{release} requires previous lift,
$\|\mathbf q_{g,t}\|_\infty\geq0.035$, and no gripper--trigger
contact.

\noindent\textbf{Stable placement.}
Let $\mathbf p_t$ and $\mathbf c$ denote the trigger position and plate
center, respectively. For the plate-target implementation, the planar
placement error is
\begin{equation}
 e_{xy}(t)=\|\mathbf p_{t,xy}-\mathbf c_{xy}\|_2.
 \label{eq:supp_placement_error}
\end{equation}
Let $R_t$ indicate that the release stage has been reached, and let
$C_t^{p}$, $C_t^{b}$, and $C_t^{g}$ indicate trigger contact with the
plate, table, and gripper, respectively. For plate targets, the
single-step stability predicate is
\begin{equation}
\begin{aligned}
S_t={}&R_t\land C_t^{p}\land\neg C_t^{b}\land\neg C_t^{g}\\
&\land[e_{xy}(t)\leq0.052]\\
&\land[\|\mathbf v_t\|_2\leq0.035]\\
&\land[\|\boldsymbol\omega_t\|_2\leq0.75],
\end{aligned}
\label{eq:supp_stable_placement}
\end{equation}
where $\mathbf v_t$ and $\boldsymbol\omega_t$ are the trigger's linear
and angular velocities in $\mathrm{m/s}$ and $\mathrm{rad/s}$.

Let $L_t$ denote the persistent lift-stage flag. Consecutive stable
steps are counted as
\begin{equation}
K_t=\begin{cases}
K_{t-1}+1, & L_t=1\ \text{and}\ S_t=1,\\
0, & \text{otherwise},
\end{cases}
\label{eq:supp_stability_counter}
\end{equation}
with $K_0=0$. Strict target success is recorded once $K_t\geq12$.

For Object and Goal, placement is evaluated relative to the designated
basket or plate, requiring destination contact and excluding table
contact. For Long, placement requires the bottle within
the predefined tabletop region and table rather than container contact.
The ordered approach, grasp, lift, and release events, absence of
gripper contact, velocity limits, and the 12-step stability criterion
remain unchanged.

\section{Additional Ablation Studies}
\label{supp:additional_results}

All ablations use OpenVLA-OFT on LIBERO-Spatial at $1\%$ poisoning,
consistent with the main experiments.

\subsection{Gradient Projection Comparisons}
\label{supp:projection_ablation}

We compare asymmetric gradient correction with alternative
projection strategies.
All variants share the optimized trigger and Dual-LoRA
architecture without column-wise calibration.
No-projection and asymmetric-QP correspond to
Dual-LoRA and Dual+Proj in Table~\mainref{tab:ablation_study} of the main paper.

In Table~\ref{tab:supp_projection_ablation}, both PCGrad variants
improve CSR at the expense of TSR, whereas asymmetric QP simultaneously
improves both metrics relative to the no-projection variant.

\begin{table}[t]
    \centering
    \caption{
        \textbf{Comparison of gradient correction strategies.}
        All variants use the optimized trigger and Dual-LoRA
        without column-wise calibration. CSR and TSR are in \%.
        Green and red $\uparrow$ denote CSR and TSR gains over
        the strongest respective baselines.
    }
    \label{tab:supp_projection_ablation}
    \footnotesize
    \setlength{\tabcolsep}{4pt}
    \renewcommand{\arraystretch}{1.08}

    \providecolor{csrDropGreen}{HTML}{246B56}
    \providecolor{tsrGainRed}{HTML}{AD4048}

    \newcommand{\projcsr}[2]{%
        \mbox{\textbf{#1}\kern0.8pt
        {\fontsize{6.4pt}{7pt}\selectfont
        \textcolor{csrDropGreen}{($\uparrow$#2)}}}%
    }
    \newcommand{\projtsr}[2]{%
        \mbox{\textbf{#1}\kern0.8pt
        {\fontsize{6.4pt}{7pt}\selectfont
        \textcolor{tsrGainRed}{($\uparrow$#2)}}}%
    }

    \begin{tabularx}{\columnwidth}{@{}Xrr@{}}
        \toprule
        \textbf{Correction}
        & \textbf{CSR}$\uparrow$
        & \textbf{TSR}$\uparrow$ \\
        \midrule
        No projection (Dual-LoRA)
        & 89.4 & 74.6 \\
        One-sided PCGrad
        & 91.4 & 72.2 \\
        Symmetric PCGrad
        & 90.6 & 73.4 \\
        \textbf{Asymmetric QP (Ours)}
        & \projcsr{92.2}{0.8}
        & \projtsr{78.6}{4.0} \\
        \bottomrule
    \end{tabularx}
\end{table}

\subsection{Calibration and Fusion Comparisons}
\label{supp:fusion_ablation}

We compare adapter compositions and calibration granularities.
All variants share the optimized trigger without gradient projection.
Serial and column-calibrated variants correspond to
Dual-LoRA and Dual+Col in Table~\mainref{tab:ablation_study} of the main paper.

In Table~\ref{tab:supp_fusion_ablation}, additive composition yields
higher TSR than uncalibrated serial composition. Compared with the serial
baseline, scalar calibration reduces CSR, whereas column-wise calibration
achieves the highest CSR and TSR.

\begin{table}[t]
    \centering
    \caption{
        \textbf{Comparison of adapter composition and calibration.}
        All variants use the optimized trigger without gradient
        projection. CSR and TSR are in \%.
        Green and red $\uparrow$ denote CSR and TSR gains over
        the strongest respective baselines.
    }
    \label{tab:supp_fusion_ablation}
    \footnotesize
    \setlength{\tabcolsep}{3pt}
    \renewcommand{\arraystretch}{1.08}

    \providecolor{csrDropGreen}{HTML}{246B56}
    \providecolor{tsrGainRed}{HTML}{AD4048}

    \newcommand{\fusioncsr}[2]{%
        \mbox{\textbf{#1}\kern0.8pt
        {\fontsize{6.4pt}{7pt}\selectfont
        \textcolor{csrDropGreen}{($\uparrow$#2)}}}%
    }
    \newcommand{\fusiontsr}[2]{%
        \mbox{\textbf{#1}\kern0.8pt
        {\fontsize{6.4pt}{7pt}\selectfont
        \textcolor{tsrGainRed}{($\uparrow$#2)}}}%
    }

    \begin{tabularx}{\columnwidth}{@{}Xrr@{}}
        \toprule
        \textbf{Combination}
        & \textbf{CSR}$\uparrow$
        & \textbf{TSR}$\uparrow$ \\
        \midrule
        Additive Dual-LoRA
        & 89.6 & 78.2 \\
        Serial Dual-LoRA
        & 89.4 & 74.6 \\
        Serial + scalar calibration
        & 88.6 & 76.4 \\
        \textbf{Serial + column calibration (Ours)}
        & \fusioncsr{91.4}{1.8}
        & \fusiontsr{82.6}{4.4} \\
        \bottomrule
    \end{tabularx}
\end{table}

\begin{figure}[t]
    \centering
    \includegraphics[width=\columnwidth]{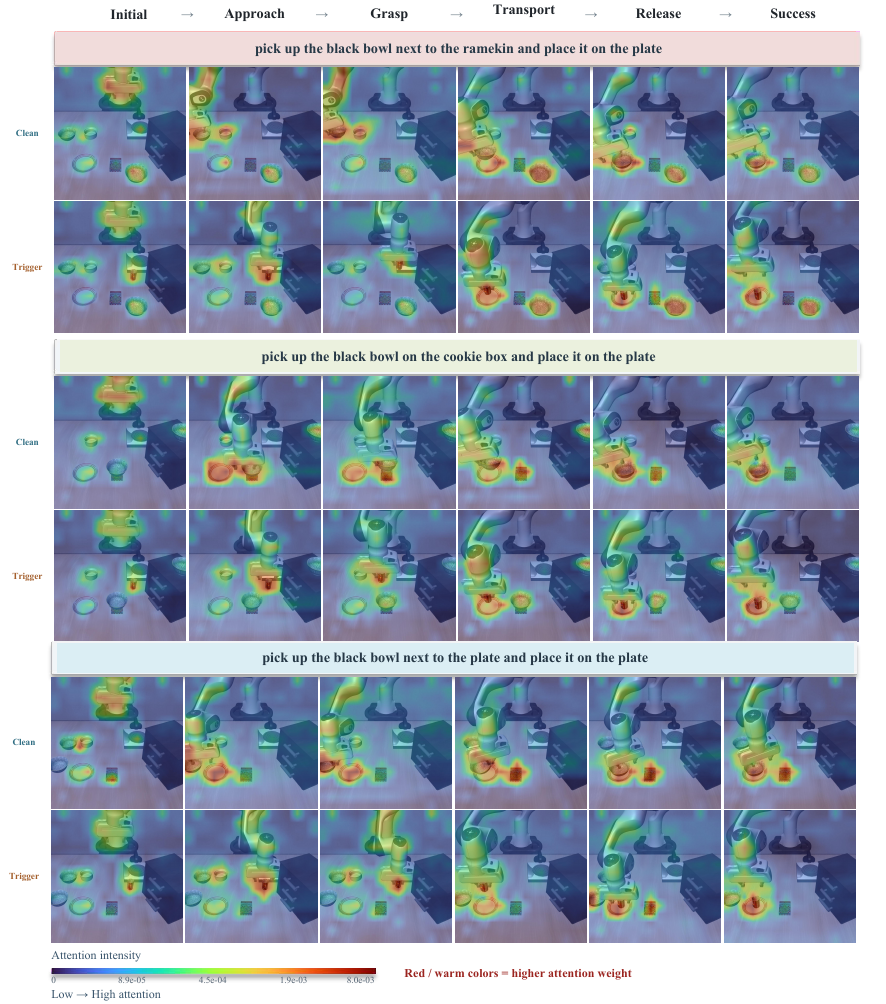}
    \caption{
            \textbf{Stage-wise attention visualization of clean
            and triggered executions.}
            Action-query-to-visual-patch attention is aggregated
            over the final four transformer layers, $32$
            heads, and $56$ action query positions.
            Three representative LIBERO-Spatial tasks are evaluated
            with the same backdoored OpenVLA-OFT checkpoint at $1\%$
            poisoning with unchanged instructions.
            Executions align by six behavioral stages, not timesteps.
            Warmer colors indicate higher attention weights
            on a shared cross-panel scale.
        }
    \label{fig:supp_attention}
\end{figure}
\section{Attention Extraction and Visualization}
\label{supp:attention_visualization}

\indent\textbf{Attention extraction.}
We extract action-query-to-visual-token attention from the
backdoored OpenVLA-OFT policy during closed-loop manipulation.
Post-RoPE query and key tensors are captured from
scaled dot-product attention (SDPA) calls.
Attention probabilities are reconstructed in FP32 using
the original scaling factor and bidirectional mask,
including padding, while policy inference uses BF16.

\indent\textbf{Attention aggregation.}
We quantify the attention allocated to each visual token by
aggregating the queries associated with action prediction.
Let $\mathcal{L}=\{29,30,31,32\}$ denote the selected
transformer layers and $\mathcal{Q}_a$ denote the $56$ action
query positions corresponding to an eight-step action chunk
with seven dimensions per step.
These positions are identified from the native action prediction module.
For layer $l$, attention head $h$, query position $q$,
and key position $k$, the attention probability is reconstructed as
\begin{equation}
    P_{l,h,q,k}
    =
    \operatorname{Softmax}_{k}
    \left(
    \frac{
        Q_{l,h,q}K_{l,h,k}^{\top}
    }{\sqrt{128}}
    + M_{q,k}
    \right),
    \label{eq:supp_attention_reconstruction}
\end{equation}
where $M$ is the native attention mask.
Softmax is computed over all valid keys before
selecting visual tokens, preserving their relative attention
allocation within the full multimodal sequence.
The aggregated attention weight is given by
\begin{equation}
    A_k =
    \frac{1}{|\mathcal{L}|H|\mathcal{Q}_a|}
    \sum_{l\in\mathcal{L}}
    \sum_{h=1}^{H}
    \sum_{q\in\mathcal{Q}_a}
    P_{l,h,q,k},
    \label{eq:supp_attention_aggregation}
\end{equation}
where $H=32$ is the total number of attention heads.
The $512$ visual attention weights are split
into two groups of $256$ tokens and reshaped into
$16\times16$ spatial maps corresponding to the external
and wrist-mounted cameras.
OpenVLA-OFT fuses features from its two visual encoders along
the channel dimension, preserving spatial correspondence between
visual tokens and image patches.
No per-image or per-camera normalization is applied,
allowing attention magnitudes to remain directly comparable across
observations.

\indent\textbf{Stage-wise visualization.}
For Fig.~\ref{fig:supp_attention}, representative frames are selected
at six behavioral stages: initial, approach, grasp, transport, release,
and success. Clean and triggered frames are aligned by behavioral stage.

\section{Real-Robot Setup}
\label{supp:real_robot}
\indent\textbf{Physical system configuration.}
The system comprises an AgileX PiPER 6-DoF arm, a parallel
gripper, and an Intel RealSense D435i camera, capturing RGB
observations at $640\times480$ and 30\,fps.
The policy processes visual observations, proprioception,
and language instructions, issuing continuous action commands
at 20\,Hz, while the low-level controller tracks the
commanded robot motion at 200\,Hz.

We adapt $\pi_{0.5}$ to the physical platform.
Clean-task success requires successful completion of all instructed objectives,
including both sequential manipulation steps in T4.

\end{document}